\documentclass[lettersize,journal]{IEEEtran}
\usepackage{amsmath,amsfonts}
\usepackage{algorithmic}
\usepackage[ruled,linesnumbered]{algorithm2e}
\usepackage{amsmath}
\usepackage{amssymb}

\usepackage{array}
\usepackage[caption=false,font=normalsize,labelfont=sf,textfont=sf]{subfig}
\usepackage{textcomp}
\usepackage{stfloats}
\usepackage[switch]{lineno} % 加载行号，并适应双栏排版

\usepackage{url}
\usepackage{verbatim}
\usepackage{graphicx}
\usepackage[nocompress]{cite}

\usepackage{booktabs}
\usepackage{multirow}
\usepackage{graphicx}

\usepackage{xcolor}
\usepackage{hyperref}
\definecolor{sciblue}{RGB}{0,0,255}

\hypersetup{
    colorlinks=true,       % 开启彩色链接，去掉难看的方框
    linkcolor=sciblue,     % 内部引用颜色（如图1、公式2等，保持深蓝色显得严谨）
    citecolor=sciblue,     % 参考文献引用颜色（如 [1], [2] 等，保持深蓝）
    urlcolor=magenta,      % 【关键修改】网页 URL 链接颜色，完美复刻你图片里的洋红色！
    breaklinks=true        % 允许长链接自动换行
}

\usepackage[table]{xcolor}

\definecolor{gold_bg}{RGB}{255, 235, 175}    % 第一名：柔和金
\definecolor{silver_bg}{RGB}{230, 240, 250}  % 第二名：清爽银蓝
\definecolor{bronze_bg}{RGB}{235, 245, 235}  % 第三名：淡雅铜绿

\usepackage{soul,color,xcolor}      % 用于设置字体颜色
\definecolor{myColor}{rgb}{1,0,0}   % 显示修订时用这行
\makeatletter
\newcommand*{\revise}{\@ifnextchar\bgroup{\revise@}{\color{myColor}}}
\newcommand*{\revise@}[1]{{\textcolor{myColor}{#1}}}
\makeatother

\begin{document}
% \linenumbers % <--- 添加这一行来真正开启行号显示

\title{SCC-Loc: A Unified Semantic Cascade Consensus Framework for UAV Thermal Geo-Localization}

\author{
        Xiaoran~Zhang$^\dagger$,\IEEEmembership{}
        Yu~Liu$^\dagger$,\IEEEmembership{}
        Jinyu~Liang,\IEEEmembership{}
        Kangqiushi~Li,\IEEEmembership{}
        Zhiwei~Huang,\IEEEmembership{}
        Huaxin~Xiao$^{*}$\IEEEmembership{}
        % \\[1em]
        % National University of Defense Technology, Changsha, China \\[0.3em]
        % $^*$corresponding author, xiaohuaxin@nudt.edu.cn
        % <-this % stops a space
% \thanks{Xiaoran Zhang, Yu Liu, Jinyu Liang, Kangqiushi Li, Zhiwei Huang, and Huaxin Xiao are with the College of Systems Engineering, National University of Defense Technology, Changsha 410073, China (Corresponding author: Huaxin Xiao; e-mail: \{zhangxiaoran, liuyu20, liangjinyu, likangqiushi20, huangzhiwei21, xiaohuaxin\}@nudt.edu.cn). 

% This work has been submitted to the IEEE for possible publication. Copyright may be transferred without notice, after which this version may no longer be accessible.
% }

% \thanks{Xiaoran Zhang, Yu Liu, Jinyu Liang, Kangqiushi Li, Zhiwei Huang, and Huaxin Xiao are with the College of Systems Engineering, National University of Defense Technology, Changsha 410073, China (\textcolor{black}{Corresponding author: Huaxin Xiao; e-mail: xiaohuaxin@nudt.edu.cn}).

% \textcolor{black}{The e-mail addresses of the other authors are: \{zhangxiaoran, liuyu20, liangjinyu, likangqiushi20, huangzhiwei21\}@nudt.edu.cn.}}

\thanks{
\textcolor{black}{Xiaoran Zhang and Yu Liu contributed equally to this work.}
\textcolor{black}{(Corresponding author: Huaxin Xiao.)}

\textcolor{black}{Xiaoran Zhang, Yu Liu, Jinyu Liang, Kangqiushi Li, Zhiwei Huang, and Huaxin Xiao are with the College of Systems Engineering, National University of Defense Technology, Changsha 410073, China.}
\textcolor{black}{(e-mail: \{zhangxiaoran, liuyu20, liangjinyu, likangqiushi20, huangzhiwei21, xiaohuaxin\}@nudt.edu.cn).}
}

}
% This work has been submitted to the IEEE for possible publication. Copyright may be transferred without notice, after which this version may no longer be accessible.
% The paper headers
\markboth{Journal of \LaTeX\ Class Files,~Vol.~14, No.~8, August~2021}%
{Shell \MakeLowercase{\textit{et al.}}: A Sample Article Using IEEEtran.cls for IEEE Journals}

% \IEEEpubid{0000--0000/00\$00.00~\copyright~2021 IEEE}
% Remember, if you use this you must call \IEEEpubidadjcol in the second
% column for its text to clear the IEEEpubid mark.

\maketitle

\begin{abstract}
Cross-modal Thermal Geo-localization (TG) provides a robust, all-weather solution for Unmanned Aerial Vehicles (UAVs) in Global Navigation Satellite System (GNSS)-denied environments. However, profound thermal-visible modality gaps introduce severe feature ambiguity, systematically corrupting conventional coarse-to-fine registration. \textcolor{black}{To address this bottleneck, we propose SCC-Loc, a unified Semantic-Cascade-Consensus localization framework.} By sharing a single DINOv2 backbone across global retrieval and MINIMA$_{\text{RoMa}}$ matching, it minimizes memory footprint and achieves zero-shot, highly accurate absolute position estimation. Specifically, we tackle modality ambiguity by introducing three cohesive components. First, we design the Semantic-Guided Viewport Alignment (SGVA) module to adaptively optimize satellite crop regions, effectively correcting initial spatial deviations. Second, we develop the Cascaded Spatial-Adaptive Texture-Structure Filtering (C-SATSF) mechanism to explicitly enforce geometric consistency, thereby \textcolor{black}{suppressing} dense cross-modal outliers. \textcolor{black}{Finally, we propose the Consensus-Driven Reliability-Aware Position Selection (CD-RAPS) strategy to determine the optimal position by integrating physically constrained pose optimization, multi-dimensional reliability evaluation, and geographic consensus.} To address data scarcity, we construct Thermal-UAV, a comprehensive dataset providing 11,890 diverse thermal queries referenced against a large-scale satellite ortho-photo and corresponding spatially aligned Digital Surface Model (DSM). Extensive experiments demonstrate that SCC-Loc establishes a new state-of-the-art, suppressing the mean localization error to 9.37 m and providing a 7.6-fold accuracy improvement within a strict 5-m threshold over the strongest baseline. Code and dataset are available at \url{https://github.com/FloralHercules/SCC-Loc}.
\end{abstract}


\section{Introduction}

\begin{figure}[!t]
  \centering
  \includegraphics[width=\linewidth]{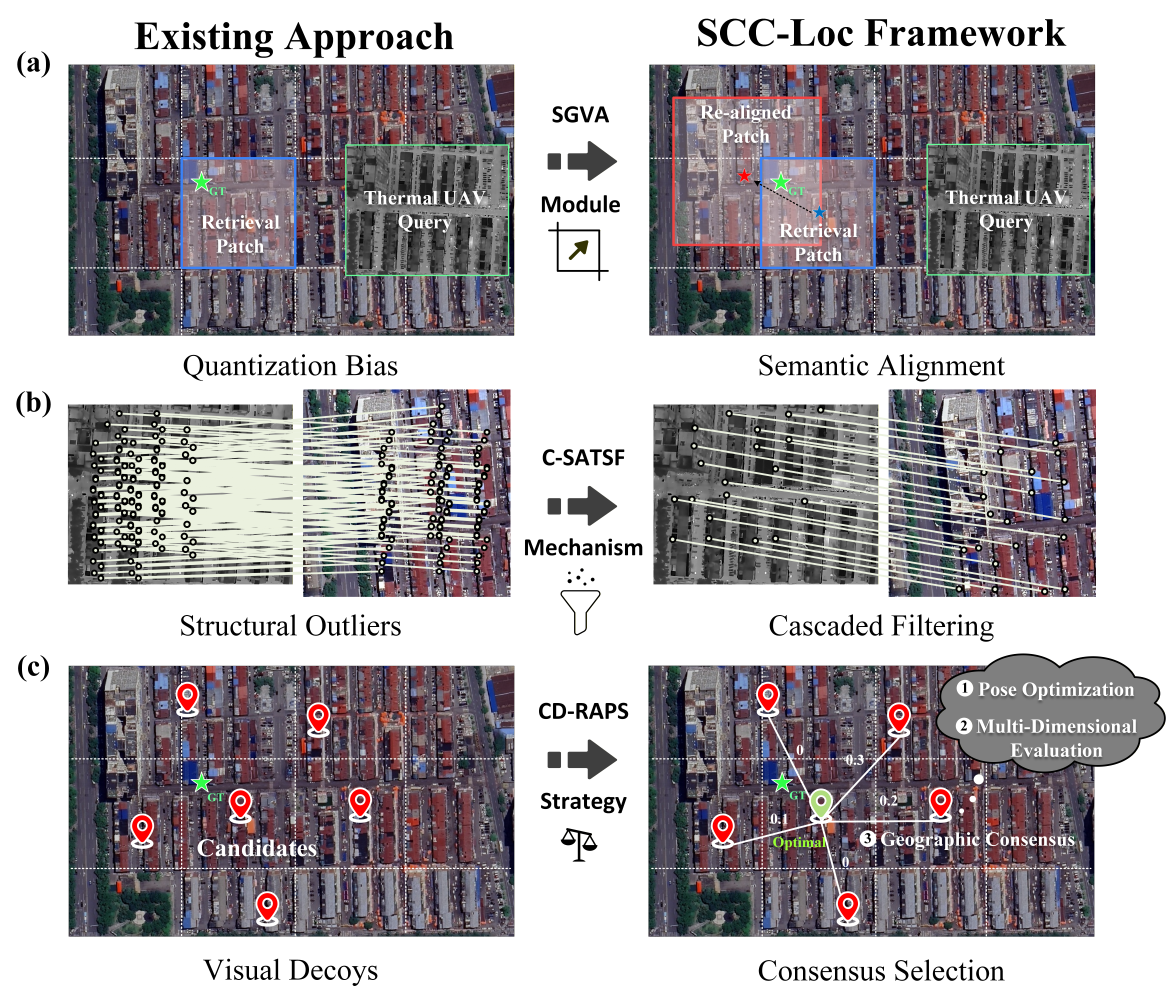}
    \caption{Conceptual comparison of the existing \textcolor{black}{approach} and our SCC-Loc framework. To overcome severe cross-modal bottlenecks in thermal geo-localization, SCC-Loc systematically resolves: (a) spatial quantization bias via semantic alignment (SGVA); \textcolor{black}{(b) dense structural outliers via cascaded filtering (C-SATSF);} and (c) deceptive visual decoys via consensus selection (CD-RAPS).}
    
    \label{Diagram}
\end{figure}

\IEEEPARstart{W}{ith} the rapid advancement of multimodal remote sensing and Earth observation technologies, Unmanned Aerial Vehicles (UAVs) have emerged as indispensable platforms for various aerial surveying and monitoring applications. However, the autonomous operation of these platforms in Global Navigation Satellite System (GNSS)-denied environments heavily relies on robust and accurate self-localization~\cite{martinez2021search, koslowski2021drones}. Visual Geo-localization (VG) has emerged as a promising solution to bridge this gap. At its core, VG determines precise geographical coordinates by establishing robust cross-view image registration and multiview matching between onboard camera imagery and geo-referenced satellite maps~\cite{zheng2020university, couturier2024review}.

Despite its tremendous potential, conventional visible-light VG is fundamentally constrained by its reliance on photometric consistency, leading to \textcolor{black}{substantial} performance degradation under severe illumination variations and nighttime conditions~\cite{xiao2023long}. To break this limitation and enable true day-and-night all-weather autonomy, cross-modal Thermal Geo-localization (TG) has recently attracted significant attention~\cite{xiao2024sthn, xiao2023long, xiao2025uasthn}. By exploiting illumination-invariant thermal infrared imagery and matching it against widely available visible-light satellite databases, TG offers a highly resilient alternative. Nevertheless, establishing reliable spatial correspondences across these highly heterogeneous domains remains a formidable cross-modal remote sensing challenge.

To achieve precise self-localization across large-scale \textcolor{black}{geo-referenced} satellite maps, state-of-the-art geo-localization \textcolor{black}{methods} almost exclusively adopt a coarse-to-fine hierarchical \textcolor{black}{framework}~\cite{he2024leveraging, ye2025exploring, meng2024airgeonet}. This standard pipeline operates by first employing global image retrieval to isolate candidate regions (coarse localization), subsequently relying on local feature matching to establish dense spatial correspondences for precise coordinate estimation (fine localization).

Despite the success of these two-stage methods on standard homologous visible-light benchmarks, shifting their application directly to cross-modal thermal-to-visible matching exposes unprecedented challenges. Because thermal radiation and solar reflectance differ fundamentally in their physical imaging mechanisms, this profound modality gap \textcolor{black}{weakens} critical local textures and structural invariants essential to conventional matching algorithms~\cite{xiao2024sthn, tuzcuouglu2024xoftr}. Consequently, direct deployment of this conventional pipeline in complex, real-world cross-modal scenarios triggers three critical bottlenecks, as conceptually illustrated in Fig.~\ref{Diagram}:

% \begin{itemize}

%     \item \textbf{Spatial quantization bias induces fatal viewport misalignments (Fig.~\ref{Diagram}(a)).} Traditional retrieval relies on databases constructed via discretized sliding windows~\cite{wu2025uav, ji2025game4loc}. Because these fixed grid centers rarely align with actual UAV positions, target regions are frequently marginalized or truncated. In texture-sparse thermal imagery, such initial spatial errors fatally deprive the subsequent fine-matching stage of sufficient overlapping regions.

%     \item \textbf{Modality discrepancies generate massive structural outliers (Fig.~\ref{Diagram}(b)).} Thermal imagery's inherent texture deficit induces severe feature ambiguity, forcing conventional matchers to overly cluster points in high-contrast areas~\cite{tuzcuouglu2024xoftr}. Furthermore, repetitive topological patterns (e.g., identical rooftops) produce widespread spurious correspondences that exhibit local structural similarity but lack global geometric rigidity. Directly utilizing these contaminated matches for pose estimation inevitably leads to catastrophic localization failures. 
    
%     \item \textbf{Visual decoys restrain robust optimal position selection (Fig.~\ref{Diagram}(c)).} In large-scale search spaces, macroscopic structural repetitions frequently yield deceptively high inlier counts for entirely incorrect locations. Existing methods merely rely on the number of inlier counts to select the best candidate pose, while falling into the ``low residual, high drift'' trap induced by these visual decoys~\cite{ye2025exploring}.
% \end{itemize}

\begin{itemize}
    \item \textcolor{black}{\textbf{Discrete satellite grids cause viewport misalignment (Fig.~1(a)).}
    In large-scale geo-localization, satellite maps are commonly divided into fixed-grid patches for retrieval. However, the actual UAV position is continuous and rarely coincides with the center of a pre-defined satellite patch. Therefore, even when the correct region is retrieved, the UAV view may be shifted toward the patch boundary or partially truncated. This practical mismatch, referred to as spatial quantization bias, reduces the valid overlap between the thermal query and the retrieved satellite candidate, thereby weakening the subsequent fine-matching stage.}

    \item \textcolor{black}{\textbf{Thermal-visible modality gaps produce dense structural outliers (Fig.~1(b)).}
    Thermal images mainly reflect radiation intensity and usually lack fine visible textures, whereas satellite maps contain rich appearance details and repetitive man-made structures. Consequently, similar rooftops, roads, and building blocks may produce many locally plausible but globally inconsistent correspondences. These unreliable matches, termed structural outliers, can severely corrupt pose estimation if they are directly used without explicit filtering.}

    \item \textcolor{black}{\textbf{Look-alike satellite regions mislead optimal position selection (Fig.~1(c)).}
    In large-scale satellite maps, different locations may share similar layouts and structural patterns. Such visually deceptive regions, termed visual decoys, may still generate internally consistent matches and produce a low reprojection error after pose optimization. However, the estimated UAV position can be far away from the true location. This phenomenon is referred to as the ``low residual, high drift'' trap, indicating that selecting the optimal pose solely based on inlier counts or fitting residuals is unreliable.}
\end{itemize}

To bypass these conventional matching bottlenecks, some pioneering thermal-specific networks (e.g., STHN~\cite{xiao2024sthn}) attempt to learn cross-modal mappings through domain-specific training. However, they heavily rely on a data-hungry supervised framework~\cite{xu2024uav, zhu2023sues, dai2023vision}. This inherently limits their zero-shot generalization capabilities, causing them to overfit to narrow scenarios and struggle when deployed in unseen environments. Moreover, they are sensitive to initial rotational misalignments against satellite maps caused by telemetry noise, severely hindering robust real-world navigation. 

Furthermore, the vulnerability of such data-hungry models is severely exacerbated by critical data scarcity. Currently, the academic community relies almost exclusively on a single open-source cross-modal dataset, \textit{Boson-nighttime}~\cite{xiao2023long}. However, this dataset exhibits severe limitations: it overly focuses on homogeneous scenes like deserts while lacking complex urban and rural topologies, and it is strictly confined to purely nighttime conditions, failing to capture diurnal thermal variations. This dual absence of spatial diversity and temporal span makes supervised cross-modal models highly prone to overfitting, impeding the comprehensive evaluation of localization algorithms in real-world environments.

To address these challenges, we propose SCC-Loc, a unified Semantic-Cascade-Consensus framework for seamless coarse-to-fine thermal geo-localization. Unlike disjointed \textcolor{black}{approaches} that suffer from error accumulation, SCC-Loc explicitly tackles the cross-modal bottlenecks through a highly cohesive pipeline: it adaptively aligns spatial viewports via semantic guidance, progressively purifies structural outliers through a cascaded sieve, and integrates physically constrained optimization, multi-dimensional evaluation, and geographic consensus voting to deduce the optimal position. Furthermore, to bridge the benchmarking gap in this domain, we construct and introduce a comprehensive Thermal-UAV dataset, providing a rigorous real-world testbed for cross-modal localization.

\textcolor{black}{The main contributions are summarized as follows:}
\begin{itemize}
    \item We propose SCC-Loc, a highly cohesive pipeline that explicitly tackles cross-modal bottlenecks and systematically bridges the semantic, structural, and geographic gaps in thermal geo-localization.

    \item We design the SGVA module. By exploiting high-level semantic activations from a shared DINOv2 backbone to adaptively optimize the crop center, it actively corrects misalignments and effectively resolves the critical spatial quantization bias.

    \item We introduce the C-SATSF mechanism. Acting as a hierarchical sieve, \textcolor{black}{it progressively enforces spatial equalization, texture verification, and structure-consistent refinement to suppress dense outliers and produce a reliable correspondence set.}

    \item We propose the CD-RAPS strategy. To overcome the \textcolor{black}{``low residual, high drift''} trap caused by visual decoys, it integrates physically constrained non-linear optimization with multi-dimensional reliability evaluation and geographic consensus voting.

    \item We construct Thermal-UAV, a diverse cross-modal dataset featuring day-night variations and complex urban-rural scenes. Extensive experiments demonstrate that SCC-Loc comprehensively outperforms existing state-of-the-art baselines on this benchmark, significantly reducing the mean localization error and establishing a new \textcolor{black}{framework} for robust all-weather UAV positioning.
\end{itemize}

The remainder of this paper is organized as follows: Section~\ref{Related Work} reviews existing literature on UAV visual geo-localization. Section~\ref{Proposed Method} details the proposed SCC-Loc framework and its core modules. Section~\ref{Experimental Setup} describes the dataset construction, evaluation metrics, baseline methods, and implementation details. Section~\ref{Results} presents the experimental results. Section~\ref{Discussion} discusses the research findings and outlines future avenues. Finally, Section~\ref{Conclusion} concludes this work.

\section{Related Work} \label{Related Work}

UAV Visual Geo-localization (VG) aims to retrieve absolute global coordinates by matching onboard imagery with geo-referenced satellite maps. Following the technological evolution, existing approaches can be categorized into cross-view image retrieval for coarse localization, fine-grained visual positioning, coarse-to-fine hierarchical frameworks, and the emerging thermal-based geo-localization.

\subsection{Cross-View Image Retrieval for Coarse Localization} 

Retrieval-based methods formulate localization as a metric learning task to search for nearest neighbors in satellite databases~\cite{avola2024uav, ji2025mmgeo, arandjelovic2016netvlad}. Early research focused on backbone evolution and semantic alignment, with Zheng et al.~\cite{zheng2020university} establishing the University-1652 benchmark. To overcome CNN receptive field limitations, FSRA~\cite{dai2021transformer} utilized Transformers for long-range dependency capture. Building upon this, CAMP~\cite{wu2024camp} introduced position-aware partitioning for layout-consistent representations, while SeGCN~\cite{liu2024segcn} leveraged Graph Convolutional Networks to explicitly model the topological relationships of semantic regions against viewpoint shifts.

Recently, Visual Foundation Models (VFMs) have introduced an off-the-shelf \textcolor{black}{framework}~\cite{keetha2023anyloc, wu2025uav, simeoni2025dinov3}. AnyLoc~\cite{keetha2023anyloc} employs self-supervised DINOv2 features with VLAD for robust zero-shot retrieval, while UAVPlace~\cite{wu2025uav} explicitly encodes rotation and scale perturbations to bolster descriptor invariance. Pushing these boundaries, DINOv3~\cite{simeoni2025dinov3} further scales representation learning for unprecedented robustness and dense semantic alignment. While effective for image-level recognition, these VFM-based methods remain limited to discrete topological outputs, failing to satisfy the demand for continuous and precise spatial coordinate estimation.

\subsection{Fine-Grained Visual Position Estimation} 

To achieve precise localization, researchers have developed methods based on feature matching and direct regression. Traditional template matching (e.g., NCC~\cite{yoo2009fast}) often fails under drastic illumination changes, whereas early deep learning-based matchers (e.g., SuperPoint~\cite{detone2018superpoint}, LoFTR~\cite{sun2021loftr}) have significantly improved robustness by establishing pixel-level 2D-2D correspondences. Building on this, more recent robust matching architectures like RoMa~\cite{edstedt2024roma} leverage dense feature correlation to handle extreme viewpoint and illumination variations. Furthermore, to specifically overcome the severe appearance variations across heterogeneous data, cross-modal matchers such as XoFTR~\cite{tuzcuouglu2024xoftr} and MINIMA~\cite{ren2025minima} have been introduced, enabling reliable feature associations even under significant modality gaps. 

Alternatively, recent approaches formulate geo-localization as an end-to-end regression task to directly predict spatial offsets. For instance, OS-FPI~\cite{chen2024fpi} employs a one-stream architecture that couples classification with offset regression for fine-grained localization. \textcolor{black}{Similarly, MMGLT~\cite{xu2026enhancing} leverages a multimodal Transformer to regress geographic coordinates within a local search region.} Although these regression-based methods yield high precision, they intrinsically rely on restrictive assumptions regarding the initial search area (i.e., requiring a known prior location) and remain highly vulnerable to large-scale cross-modal scene variations.

To balance search scope with localization precision, the coarse-to-fine two-stage strategy has become a widely adopted solution for UAV geo-localization.

\subsection{Coarse-to-Fine Hierarchical Localization} 
To balance search scope with localization precision, the coarse-to-fine two-stage strategy has become \textcolor{black}{a widely adopted solution for UAV geo-localization}~\cite{he2024leveraging}. Typical frameworks, such as AirGeoNet~\cite{meng2024airgeonet} and AnyVisLoc~\cite{ye2025exploring}, adopt a modular coupling design: candidate satellite patches are first retrieved, followed by fine coordinate solving within these regions using local feature matching or semantic alignment. For instance, Meng et al.~\cite{meng2024airgeonet} achieved robust 3-DOF pose estimation by performing dense convolutional matching between aerial features and rasterized 2D semantic maps. Furthermore, Ye et al.~\cite{ye2025exploring} established a unified benchmark integrating image retrieval, pixel-level matching, and Digital Surface Model (DSM)-based Perspective-n-Point (PnP) solving. However, existing two-stage methods often suffer from a disjoint design. The coarse and fine stages typically rely on independent feature extractors, leading to \textcolor{black}{doubled memory consumption} and computational redundancy.

\subsection{Thermal-Based UAV Geo-localization} 
While the aforementioned methods have achieved remarkable progress in RGB imagery, they inevitably experience severe performance degradation under illumination-deprived conditions (e.g., nighttime, dense fog, or smog). Consequently, leveraging thermal infrared sensors has emerged as a crucial alternative. Nevertheless, thermal-based absolute visual localization is fundamentally challenged by a severe modality gap: thermal images reflect temperature radiation (characterized by low texture and high self-similarity), whereas satellite maps depict visible light reflectance~\cite{vasile2025uav, liu2024mcgs}.

% Pioneering this field, Xiao et al.~\cite{xiao2023long} constructed the Boson-nighttime dataset—a foundational benchmark predominantly featuring homogeneous desert environments captured exclusively at night. Based on this dataset, they proposed a dual-module framework that utilizes a Thermal Generative Module (TGM) via Pix2Pix to synthesize pseudo-thermal images, followed by a Satellite-Thermal Geo-localization Module (SGM) for retrieval. To push beyond coarse retrieval, they subsequently introduced STHN~\cite{xiao2024sthn}, a deep homography estimation network that progressively regresses corner displacements for thermal-satellite alignment. This was later augmented in UASTHN~\cite{xiao2025uasthn} by integrating Crop-based Test-Time Augmentation (CropTTA) to reject low-confidence matches through uncertainty awareness. In a parallel effort, Liu et al.~\cite{liu2024multimodal} designed NIVnet, which disentangles cross-modal features into a shared shape subspace and specific attribute subspaces, thereby neutralizing modality discrepancies at the representation level.
Pioneering this field, Xiao et al.~\cite{xiao2023long} introduced the Boson-nighttime dataset—a benchmark predominantly featuring homogeneous nighttime desert environments. They proposed a dual-module framework utilizing a Thermal Generative Module (TGM) to synthesize pseudo-thermal images for subsequent Satellite-Thermal Geo-localization Module (SGM) retrieval. To advance beyond coarse retrieval, they subsequently developed STHN~\cite{xiao2024sthn}, a deep homography network for thermal-satellite alignment. This was later augmented by UASTHN~\cite{xiao2025uasthn}, utilizing uncertainty-aware Crop-based Test-Time Augmentation (CropTTA) to reject unreliable matches. Concurrently, Liu et al.~\cite{liu2024multimodal} designed NIVnet, neutralizing modality discrepancies by disentangling features into shared shape and specific attribute subspaces.

\textbf{Limitations \& Our Motivation:} Despite these trailblazing efforts, current thermal-based approaches heavily rely on domain-specific training using paired thermal-satellite data. This data-hungry \textcolor{black}{training strategy} inherently limits their \textit{generalization} capabilities, causing them to overfit to narrow scenarios and struggle when deployed in unseen environments. This vulnerability is further exacerbated by the scarcity of diverse thermal benchmarks, \textcolor{black}{as existing datasets primarily cover single regions (e.g., deserts) and narrow timeframes.}

To break this bottleneck, we tackle these limitations from both data and algorithmic perspectives. First, to address the critical data scarcity, we introduce \textbf{Thermal-UAV}, \textcolor{black}{a city-scale, multi-scene, and multi-temporal thermal-satellite localization benchmark.} Second, algorithmically, we propose \textbf{SCC-Loc}, a \textit{training-free} framework that achieves exceptional zero-shot generalization. Instead of training cross-modal feature extractors from scratch, SCC-Loc adopts a coarse-to-fine two-stage strategy that elegantly integrates off-the-shelf visual retrieval and matching models. By designing bespoke intermediate modules to bridge these heterogeneous models, we construct a unified and seamless pipeline. Evaluated comprehensively on our Thermal-UAV dataset, SCC-Loc robustly overcomes the severe cross-modal gap \textcolor{black}{without domain-specific fine-tuning}, \textcolor{black}{providing a practical framework for} highly generalizable all-weather UAV navigation.

\section{Proposed Method} \label{Proposed Method}

% \begin{figure*}[!t]
%   \centering
%   \includegraphics[width=1\linewidth]{Images/Pipeline.png}
%   \caption{Overview of the proposed SCC-Loc framework. The pipeline consists of four main stages: 
%  \textbf{(1) Feature Extraction:} A shared DINOv2 backbone extracts dense spatial features and global descriptors from the UAV thermal query and satellite database, providing modality-robust representations for subsequent retrieval and matching;
%  \textbf{(2) SGVA Module:} The Semantic-Guided Viewport Alignment module leverages the query's global token to adaptively crop and align the retrieved satellite candidates with the UAV field-of-view;
%  \textbf{(3) C-SATSF Mechanism}: Following dense matching via MINIMA$_{\text{RoMa}}$, the Cascaded Spatial-Adaptive Texture-Structure Filtering progressively purifies raw correspondences by eliminating spatial, textural, and structural inconsistencies;
%  \textbf{(4) CD-RAPS Strategy}: The Consensus-Driven Reliability-Aware Position Selection integrates 3D pose optimization and multi-dimensional reliability evaluation to vote for the robust optimal horizontal position via geographic consensus.}
%   \label{pipeline}
% \end{figure*}
\begin{figure*}[!t]
  \centering
  \includegraphics[width=1\linewidth]{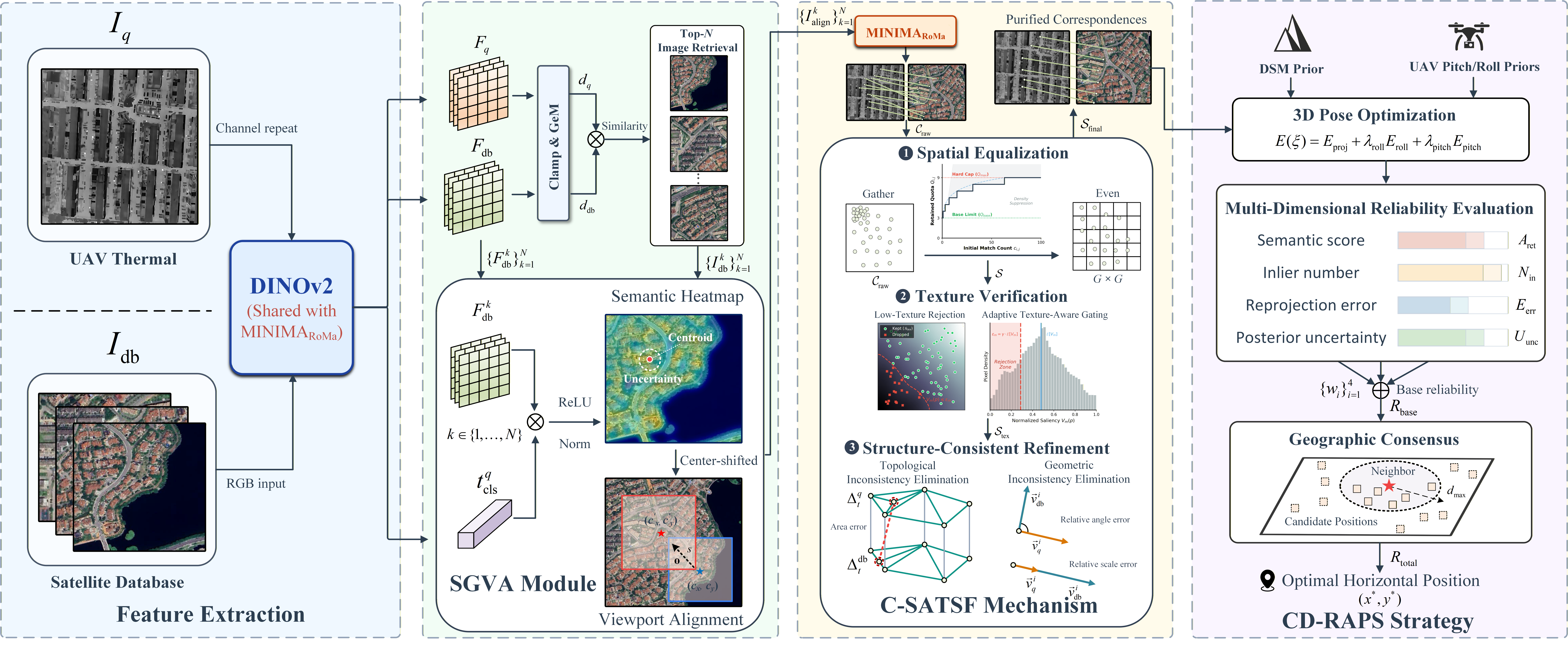}
  \caption{\textcolor{black}{Overview of the proposed SCC-Loc framework: \textbf{(1)} shared DINOv2 feature extraction for UAV thermal and satellite images; \textbf{(2)} SGVA-based viewport alignment; \textbf{(3)} C-SATSF-based correspondence purification after MINIMA$_{\text{RoMa}}$ matching; and \textbf{(4)} CD-RAPS-based position selection via pose optimization, reliability evaluation, and geographic consensus.}}
  \label{pipeline}
\end{figure*}

\subsection{Overview}
As illustrated in Fig.~\ref{pipeline}, SCC-Loc follows a cohesive coarse-to-fine \textcolor{black}{pipeline}. By sharing the DINOv2~\cite{oquab2023dinov2} backbone of the MINIMA$_{\text{RoMa}}$~\cite{ren2025minima} matcher across global retrieval and dense matching, it efficiently \textcolor{black}{mitigates the thermal-visible modality gap} without redundant memory overhead. First, the Semantic-Guided Viewport Alignment (SGVA) module corrects spatial quantization bias by aligning the viewport center via the UAV's global [CLS] token. Next, the aligned patches undergo dense matching, where the Cascaded Spatial-Adaptive Texture-Structure Filtering (C-SATSF) acts as a hierarchical sieve to distill dependable correspondences and reject massive cross-modal outliers. \textcolor{black}{Finally, to avoid the ``low residual, high drift'' trap induced by visual decoys, the Consensus-Driven Reliability-Aware Position Selection (CD-RAPS) utilizes physically constrained non-linear optimization, multi-dimensional reliability evaluation, and geographic consensus to robustly determine the final position.}

\subsection{Semantic-Guided Viewport Alignment Module}

\textcolor{black}{
To rectify the spatial misalignment between the Top-$N$ retrieved satellite candidates and the thermal UAV field of view, we propose the SGVA module. This module exploits high-level semantic activations extracted by the shared DINOv2 backbone to estimate the semantic centroid of each satellite candidate, and then adjusts its crop center with a moderate scale expansion. In this way, SGVA improves the viewport alignment between satellite imagery and the thermal query for subsequent fine-grained matching.
}

Formally, the global retrieval process begins by feeding the thermal query $I_q$ and the visible database images $I_{\text{db}}$ into the shared DINOv2 backbone $\Phi_{\text{DINOv2}}$ to extract their corresponding dense feature maps, denoted as $F_q \in \mathbb{R}^{H \times W \times D}$ and $F_{\text{db}} \in \mathbb{R}^{H \times W \times D}$, respectively. To derive a compact global representation, we treat the dense feature map for each modality $m \in \{q, \text{db}\}$ as a set of spatial feature tokens $\mathcal{X}_m$, where the cardinality is $|\mathcal{X}_m| = H \times W$. These tokens are then aggregated via Generalized Mean (GeM) pooling~\cite{radenovic2018fine} to obtain the modality-invariant global descriptor $d_m \in \mathbb{R}^D$:

\begin{equation}
    d_m = \left( \frac{1}{|\mathcal{X}_m|} \sum_{x \in \mathcal{X}_m} \operatorname{clamp}(x, \epsilon_{\text{min}})^\psi \right)^{\frac{1}{\psi}}, \quad m \in \{q, \text{db}\},
\end{equation}
where $x \in \mathcal{X}_m$ represents an individual spatial token of dimension $D$. The pooling exponent $\psi$ is configured as a fixed empirical constant, while $\operatorname{clamp}(\cdot, \epsilon_{\text{min}})$ enforces a lower bound $\epsilon_{\text{min}}$ to discard negative and negligible activations, \textcolor{black}{thereby encouraging the pooling mechanism to focus on high-response regions.} Following the $L_2$-normalization of the resulting pooled descriptors $d_q$ and $d_{\text{db}}$, we compute their cosine similarity and rank the database images to retrieve the top-$N$ candidate set $\mathcal{R} = \{I_{\text{db}}^k\}_{k=1}^N$~\cite{arandjelovic2016netvlad}.

Having obtained the coarse candidate set $\mathcal{R}$, we next establish a semantic interaction mechanism to \textcolor{black}{adaptively adjust the satellite viewport} for each $I_{\text{db}}^k \in \mathcal{R}$. \textcolor{black}{Unlike the retrieval-oriented GeM descriptor used for global ranking,} we \textcolor{black}{use} the [CLS] token $t_{\text{cls}}^{q} \in \mathbb{R}^D$~\cite{dosovitskiy2020image}, \textcolor{black}{extracted by} $\Phi_{\text{DINOv2}}$, to \textcolor{black}{represent} the global semantic context of the UAV view, and \textcolor{black}{correlate it with} the dense patch features $F_{\text{db}}^k \in \mathbb{R}^{H \times W \times D}$ of the $k$-th satellite candidate. This interaction generates a semantic correlation heatmap $M \in \mathbb{R}^{H \times W}$ via pixel-wise cosine similarity, providing precise spatial guidance for the subsequent adaptive cropping:

\begin{equation}
    M_{i,j} = \langle \text{Norm}(t_{\text{cls}}^{q}), \text{Norm}(F_{\text{db}}^k[i,j]) \rangle,
\end{equation}
where $(i,j)$ denotes the spatial coordinates within the $H \times W$ feature map, $\text{Norm}(\cdot)$ denotes the $L_2$-normalization operation, and $\langle \cdot, \cdot \rangle$ represents the dot product. This heatmap highlights regions in the satellite imagery that semantically align with the global view of the UAV.

To distill precise geometric cues from $M$, we interpret it as a probability distribution. Specifically, we apply ReLU rectification and normalization to obtain the spatial probability mass function $\mathbf{P}$:

\begin{equation}
    \mathbf{P}_{i,j} = \frac{\text{ReLU}(M_{i,j})}{\sum_{u,v} \text{ReLU}(M_{u,v})}.
\end{equation}
% where $(i,j)$ denotes the spatial grid indices.

\textcolor{black}{We then compute the center of gravity $\mu_{\text{map}}$ from the spatial probability distribution $\mathbf{P}$ to estimate the semantic target centroid:}

% \begin{equation}
%     \mu_{\text{map}} = \sum_{i,j} \mathbf{P}_{i,j} \cdot \mathbf{x}_{i,j}, \quad \sigma_{\text{map}} = \sqrt{\sum_{i,j} \mathbf{P}_{i,j} \cdot \| \mathbf{x}_{i,j} - \mu_{\text{map}} \|^2},
% \end{equation}

\begin{equation}
{\color{black}
    \mu_{\text{map}} = \sum_{i,j} \mathbf{P}_{i,j} \cdot \mathbf{x}_{i,j},
}
\end{equation}
\textcolor{black}{where $\mathbf{x}_{i,j}$ represents the 2D normalized spatial coordinates corresponding to the grid index $(i, j)$.}

\textcolor{black}{
Based on the estimated semantic centroid $\mu_{\text{map}}$ and the center of the original satellite tile, we introduce a lightweight viewport adaptation mechanism to derive the aligned crop parameters, including the normalized center shift $\mathbf{o}=(o_x,o_y)$ and the scale factor $s$. To reduce empirical parameter dependency and avoid excessive viewport deformation, the aligned viewport parameters are calculated as follows:
}

% \begin{equation}
% \left\{
% \begin{aligned}
%     \eta &= \lambda \cdot \sigma_{\text{map}}, \\
%     g &= 1 + \alpha (1 - \eta), \\
%     s &= 1 + \beta \cdot \eta, \\
%     \mathbf{o} &= (\mu_{\text{map}} - 0.5) \cdot g,
% \end{aligned}
% \right.
% \end{equation}

\begin{equation}
{\color{black}
\left\{
\begin{aligned}
    s &= 1 + \alpha, \\
    \mathbf{o} &= (\mu_{\text{map}} - 0.5) \cdot \beta,
\end{aligned}
\right.
}
\end{equation}
\textcolor{black}{where $\alpha$ controls the expansion rate of the satellite viewport, and $\beta$ controls the strength of the semantic center shift. The constant $0.5$ denotes the normalized center of the original satellite tile. In this formulation, the crop center is shifted toward the semantic centroid $\mu_{\text{map}}$, while the crop scale is moderately enlarged to preserve sufficient contextual information for subsequent dense matching.}

\textcolor{black}{Subsequently, given the initial satellite retrieval patch with physical dimensions $w \times h$ and center $(c_x,c_y)$, the geometry of the newly adapted crop, defined by its updated center $(c'_x,c'_y)$ and dimensions $(w',h')$, is formulated as:}

\begin{equation}
    \begin{aligned}
        c'_x &= c_x + o_x \cdot w, \quad & w' &= s \cdot w, \\
        c'_y &= c_y + o_y \cdot h, \quad & h' &= s \cdot h.
    \end{aligned}
\end{equation}

\textcolor{black}{
Intuitively, this formulation decouples semantic center correction from contextual preservation. The semantic centroid $\mu_{\text{map}}$ determines the direction of viewport relocation, while $\beta$ controls the magnitude of the center shift to avoid insufficient or excessive displacement. Meanwhile, $\alpha$ enlarges the field of view to retain peripheral context and reduce the risk of truncating the true UAV-visible region. This simplified SGVA design reduces empirical parameter dependency and provides a stable trade-off between semantic center alignment and contextual preservation.
}

Finally, based on these updated geometric parameters, we extract the adaptively aligned satellite viewport, denoted as $I_{\text{align}}^k$. This aligned patch $I_{\text{align}}^k$, along with the thermal UAV query $I_q$, \textcolor{black}{is forwarded} to the MINIMA$_{\text{RoMa}}$ model for fine-grained matching.

\textcolor{black}{By explicitly resolving region-level spatial misalignment, SGVA provides better-aligned satellite patches for MINIMA$_{\text{RoMa}}$, enabling the matcher to establish more reliable dense cross-spectral correspondences.}

The complete workflow of the SGVA module is shown in Algorithm~\ref{alg:SGVA}.

\begin{algorithm}[!t]
\DontPrintSemicolon
\caption{Semantic-Guided Viewport Alignment Module}
\label{alg:SGVA}

% 定义带有左侧竖线包裹的自定义块
\SetKwBlock{PartOne}{\textbf{Part 1: Semantic Activation}}{}
\SetKwBlock{PartTwo}{\textbf{Part 2: Parameter Calculation}}{}
\SetKwBlock{PartThree}{\textbf{Part 3: Viewport Alignment}}{}

\SetKwInOut{Input}{Input}
\SetKwInOut{Output}{Output}

\Input{
  UAV query semantics $t_{\text{cls}}^q$; Satellite features $F_{\text{db}}^k$; 
  Candidate image $I_{\text{db}}^k$ with center $(c_x, c_y)$ and size $(w, h)$.
}
\Output{
  Aligned satellite viewport $I_{\text{align}}^k$.
}

\BlankLine
\PartOne{
    $M \leftarrow \text{CosineSimilarity}(t_{\text{cls}}^q, F_{\text{db}}^k)$\;
    $\mathbf{P} \leftarrow \text{Normalize}(\text{ReLU}(M))$\;
}

\BlankLine
% \PartTwo{
%     $\mu_{\text{map}}, \sigma_{\text{map}} \leftarrow \text{EstimateDistribution}(\mathbf{P})$\;
%     $\eta \leftarrow \lambda \cdot \sigma_{\text{map}}$\;
%     $g \leftarrow 1 + \alpha(1 - \eta)$\;
%     $s \leftarrow 1 + \beta \cdot \eta$\;
%     $\mathbf{o} \leftarrow (\mu_{\text{map}} - 0.5) \cdot g$\;
% }
\PartTwo{ 
{\color{black}
    $\mu_{\text{map}} \leftarrow \text{EstimateCentroid}(P)$\;
    $s \leftarrow 1 + \alpha$\;
    $\mathbf{o} \leftarrow (\mu_{\text{map}} - 0.5) \cdot \beta$\;
}
}
\BlankLine
\PartThree{
    $(c_x', c_y') \leftarrow (c_x, c_y) + \mathbf{o} \cdot (w, h)$\;
    $I_{\text{align}}^k \leftarrow \text{Crop}(I_{\text{db}}^k, c_x', c_y', s \cdot w, s \cdot h)$\;
}

\textbf{return} {$I_{\text{align}}^k$}\;

\end{algorithm}

\subsection{Cascaded Spatial-Adaptive Texture-Structure Filtering Mechanism}

% To purify the dense tentative correspondences established by the $\text{MINIMA}_{\text{RoMa}}$ matcher into a coherent set of highly reliable inliers, effectively suppressing the substantial outliers induced by severe cross-modal discrepancies, we propose the C-SATSF mechanism. C-SATSF functions as a hierarchical sieve, systematically enforcing constraints through three progressive operations: spatial equalization, texture verification, and structure-consistent refinement.

\textcolor{black}{To obtain a coherent set of reliable inliers from the dense tentative correspondences generated by the $\text{MINIMA}_{\text{RoMa}}$ matcher, we propose the C-SATSF mechanism. It progressively suppresses outliers induced by severe cross-modal discrepancies through three operations: spatial equalization, texture verification, and structure-consistent refinement.}

\subsubsection{Density-Aware Spatial Equalization} 
Conventional confidence-based filtering tends to cluster keypoints in high-contrast regions (e.g., isolated prominent structures) while neglecting sparse but geometrically critical areas. This uneven spatial distribution leads to ill-conditioned configurations for the subsequent PnP solver~\cite{wu2006pnp}.

% To counteract this, we employ a grid-based regularization with a logarithmic dynamic quota. We partition the image domain into a $G \times G$ grid. For each cell $(i,j)$, instead of enforcing a rigid capacity, we retain a dynamic number of matches $Q_{i,j}$ derived from the local feature density. Given the initial raw match set $\mathcal{C}_{\text{raw}}$, let $\mathcal{C}_{i,j}$ denote the subset of raw matches falling within cell $(i,j)$, and $c_{i,j} = |\mathcal{C}_{i,j}|$ be its cardinality. The dynamic quota and the spatially equalized output set $\mathcal{S}$ are jointly formulated as:

To \textcolor{black}{alleviate this spatial imbalance}, we employ a grid-based regularization strategy with a logarithmic dynamic quota. Specifically, we partition the image domain into a $G \times G$ grid and \textcolor{black}{adaptively retain matches in each cell according to its local feature density}. For each cell $(i,j)$, instead of using a \textcolor{black}{fixed retention quota}, we assign a cell-specific quota $Q_{i,j}$. Given the initial raw match set $\mathcal{C}_{\text{raw}}$, let $\mathcal{C}_{i,j}$ denote the subset of raw matches located in cell $(i,j)$, and let $c_{i,j}=|\mathcal{C}_{i,j}|$ be its cardinality. The dynamic quota and the spatially equalized output set $\mathcal{S}$ are jointly formulated as:

\begin{equation}
\left\{
\begin{aligned}
    Q_{i,j} &= \min \left( Q_{\text{base}} + \lfloor \log_{2}(c_{i,j}+1) \rfloor, \, Q_{\text{max}} \right), \\
    \mathcal{S} &= \bigcup_{i,j=1}^{G} \operatorname{TopK}(\mathcal{C}_{i,j}, \, Q_{i,j}),
\end{aligned}
\right.
\end{equation}
where $Q_{\text{base}}$ is the minimum reservation limit, $Q_{\text{max}}$ is the absolute upper bound (empirically set to $3 Q_{\text{base}}$), $\lfloor \cdot \rfloor$ denotes the floor operation, $\operatorname{TopK}(\cdot, n)$ extracts the top $n$ matches based on their initial confidence scores, and $\bigcup_{i,j=1}^{G}$ denotes the set union across all $G \times G$ spatial cells.

This logarithmic quota \textcolor{black}{allows texture-rich regions to contribute slightly more reliable matches}, while the upper bound $Q_{\text{max}}$ prevents them from \textcolor{black}{overwhelming} the global match distribution. \textcolor{black}{In this way, the retained correspondences maintain more balanced spatial coverage}, providing a robust initial set $\mathcal{S}$ for the subsequent texture verification stage.

\subsubsection{Adaptive Texture Saliency Verification}

Following spatial equalization, we address the matching ambiguity prevalent in low-texture regions—a challenge \textcolor{black}{further aggravated} by the modality gap between UAV thermal imagery (which often exhibits low contrast) and visible satellite imagery (which provides abundant high-frequency details).

We first compute a local saliency map $V_m$ for each modality $m \in \{q, \text{db}\}$ to quantify texture richness. For a given keypoint $p$, let $\mathcal{W}(p)$ denote its spatial neighborhood. The local variance $\sigma_m^2(p)$ within $\mathcal{W}(p)$ and the subsequent normalized local saliency $V_m(p)$ are jointly formulated as:

\begin{equation}
\left\{
\begin{aligned}
    \sigma_m^2(p) &= \mathbb{E}[I_m^2(p)] - \left( \mathbb{E}[I_m(p)] \right)^2, \\
    V_m(p) &= \frac{\sigma_m(p) - \sigma_{m}^{\min}}{\sigma_{m}^{\max} - \sigma_{m}^{\min}},
\end{aligned}
\right.
\end{equation}
where $\mathbb{E}[\cdot]$ denotes the expectation operator, $I_m(p)$ represents the normalized image intensity, and $\sigma_{m}^{\min}$ and $\sigma_{m}^{\max}$ denote the global minimum and maximum standard deviations across the respective modality. This effectively maps the local texture richness to a unified saliency scale $V_m(p) \in [0, 1]$.

To actively reject unreliable matches in feature-poor areas (e.g., large water bodies), we establish an adaptive saliency threshold $\epsilon_m$ for each modality. Given the spatially equalized match set $\mathcal{S}$ consisting of corresponding keypoint pairs $(p_q, p_{\text{db}})$, the texture-aware gating mechanism formulates this threshold and outputs a purified subset $\mathcal{S}_{\text{tex}}$ as follows:

\begin{equation}
\left\{
\begin{aligned}
    \epsilon_m &= \gamma \cdot \mathbb{E}[V_m], \quad m \in \{q, \text{db}\}, \\
    \mathcal{S}_{\text{tex}} &= \left\{ (p_q, p_{\text{db}}) \in \mathcal{S} \mid V_q(p_q) > \epsilon_q \land V_{\text{db}}(p_{\text{db}}) > \epsilon_{\text{db}} \right\},
\end{aligned}
\right.
\end{equation}
where $\gamma$ is a threshold scaling factor, and $\land$ denotes the logical AND operator, strictly requiring a match to surpass the adaptive saliency lower bound in both modalities simultaneously. By effectively pruning low-texture ambiguous matching pairs, this filtering layer significantly reduces the outlier ratio, ensuring that only correspondences anchored on salient physical structures are forwarded to the subsequent geometric refinement stage.
\subsubsection{Structure-Consistent Geometric Refinement}

Following texture saliency verification, the surviving correspondences may still contain spatially erroneous outliers caused by semantically similar but repetitive patterns (e.g., identical rooftops). To eliminate these structural outliers, this final geometric refinement stage enforces structural consistency by unifying local topological invariance and global geometric rigidity.

\textcolor{black}{We leverage Delaunay triangulation to construct a local spatial graph over the UAV keypoints~\cite{ma2019locality, jiang2019reliable}. Since the local transformation between the UAV view and the satellite ortho-photo can be approximated by a similarity transformation, the area ratio of corresponding triangles should remain approximately invariant across modalities.} Let $\Delta_t^q$ and $\Delta_t^{\text{db}}$ denote the $t$-th triangle pair in the query and database images, respectively. Based on this assumption, the cross-modal area ratio $\rho_t$ and its topological deviation $\delta_t$ from the global consensus are formulated as:

\begin{equation}
\left\{
\begin{aligned}
    \rho_t &= \frac{\operatorname{Area}(\Delta_t^{\text{db}})}{\operatorname{Area}(\Delta_t^{q})}, \\
    \delta_t &= \frac{\left| \rho_t - \operatorname{med}(\boldsymbol{\rho}) \right|}{\operatorname{med}(\boldsymbol{\rho})},
\end{aligned}
\right.
\end{equation}
where $\operatorname{Area}(\cdot)$ computes the geometric area of a given triangle, $\boldsymbol{\rho}$ denotes the set of all triangle area ratios, and $\operatorname{med}(\cdot)$ represents the robust median operator. 

% To gracefully handle local distortions, we deploy a local voting mechanism on the texture-filtered set $\mathcal{S}_{\text{tex}}$. Let $\mathcal{T}(p_q)$ denote the local neighborhood set of triangles sharing the query keypoint $p_q$, and $\Delta_t^q \in \mathcal{T}(p_q)$ represent the $t$-th connected triangle in the UAV thermal view. A triangle is penalized if its topological deviation $\delta_t$ exceeds a topological tolerance threshold $\epsilon_{\text{topo}}$. The negative vote rate $\kappa(p_q)$ and the resulting topologically purified subset $\mathcal{S}_{\text{topo}}$ are sequentially formulated as:

To \textcolor{black}{account for} local distortions, we \textcolor{black}{introduce} a local voting mechanism on the texture-filtered set $\mathcal{S}_{\text{tex}}$. Let $\mathcal{T}(p_q)$ denote the \textcolor{black}{set of neighboring triangles that share} the query keypoint $p_q$, and \textcolor{black}{let} $\Delta_t^q \in \mathcal{T}(p_q)$ represent the $t$-th connected triangle in the UAV thermal view. A triangle is \textcolor{black}{regarded as inconsistent} if its topological deviation $\delta_t$ exceeds the tolerance threshold $\epsilon_{\text{topo}}$. The negative vote rate $\kappa(p_q)$ and the resulting \textcolor{black}{topologically refined} subset $\mathcal{S}_{\text{topo}}$ are then formulated as:

\begin{equation}
\left\{
\begin{aligned}
    \kappa(p_q) &= \frac{1}{|\mathcal{T}(p_q)|} \sum_{\Delta_t^q \in \mathcal{T}(p_q)} \mathbb{I}(\delta_t > \epsilon_{\text{topo}}), \\
    \mathcal{S}_{\text{topo}} &= \left\{ (p_q, p_{\text{db}}) \in \mathcal{S}_{\text{tex}} \mid \kappa(p_q) \le 0.5 \right\},
\end{aligned}
\right.
\end{equation}
where $\mathbb{I}(\cdot)$ denotes the indicator function, and $|\mathcal{T}(p_q)|$ is the number of \textcolor{black}{neighboring triangles incident to} $p_q$. Matches with a negative vote rate greater than $50\%$ are \textcolor{black}{regarded as} topological anomalies and \textcolor{black}{removed} from the candidate pool.

\begin{algorithm}[!t]
\DontPrintSemicolon
\caption{Cascaded Spatial-Adaptive Texture-Structure Filtering Mechanism}
\label{alg:CSATSF}

% 定义带有左侧竖线包裹的自定义块
\SetKwBlock{PartOne}{\textbf{Part 1: Density-Aware Spatial Equalization}}{}
\SetKwBlock{PartTwo}{\textbf{Part 2: Adaptive Texture Saliency Verification}}{}
\SetKwBlock{PartThree}{\textbf{Part 3: Structure-Consistent Geometric Refinement}}{}

\SetKwInOut{Input}{Input}
\SetKwInOut{Output}{Output}

\Input{
  UAV Thermal Query $I_q$; Aligned Satellite Viewport $I_{\text{align}}^k$.
}
\Output{
  \textcolor{black}{Reliable} cross-modal inlier set $\mathcal{S}_{\text{final}}^{(k)}$.
}

\BlankLine
$\mathcal{C}_{\text{raw}} \leftarrow \text{MINIMA}_{\text{RoMa}}(I_q, I_{\text{align}}^k)$\;

\BlankLine
\PartOne{
    $\mathcal{Q} \leftarrow \text{ComputeLogQuotas}(\mathcal{C}_{\text{raw}})$\;
    $\mathcal{S} \leftarrow \text{SpatialEqualize}(\mathcal{C}_{\text{raw}}, \mathcal{Q})$\;
}

\BlankLine
\PartTwo{
    $V_q, V_{\text{db}} \leftarrow \text{ComputeLocalSaliency}(\mathcal{S})$\;
    $\epsilon_q, \epsilon_{\text{db}} \leftarrow \gamma \cdot \mathbb{E}[V_q], \gamma \cdot \mathbb{E}[V_{\text{db}}]$\;
    $\mathcal{S}_{\text{tex}} \leftarrow \{(p_q, p_{\text{db}}) \in \mathcal{S} \mid V_q(p_q) > \epsilon_q \land V_{\text{db}}(p_{\text{db}}) > \epsilon_{\text{db}} \}$\;
}

\BlankLine
\PartThree{
    $\kappa(p_q) \leftarrow \text{ComputeNegativeVoteRate}(\mathcal{S}_{\text{tex}})$\;
    $\mathcal{S}_{\text{topo}} \leftarrow \{(p_q, p_{\text{db}}) \in \mathcal{S}_{\text{tex}} \mid \kappa(p_q) \le 0.5 \}$\;
    $\mathcal{S}_{\text{final}}^{(k)} \leftarrow \text{EnforceGlobalConsistency}(\mathcal{S}_{\text{topo}}, \epsilon_{\text{ang}}, \epsilon_{\text{scale}})$\;
}

\textbf{return} {$\mathcal{S}_{\text{final}}^{(k)}$}\;

\end{algorithm}

% Complementarily, we enforce global consistency in scale and orientation on the surviving set $\mathcal{S}_{\text{topo}}$ to prevent structural collapse. Let $\bar{p}_q$ and $\bar{p}_{\text{db}}$ denote the geometric centroids of the keypoints in $\mathcal{S}_{\text{topo}}$ for the query and database modalities, respectively. We define the heading vectors $\vec{v}_q^i$ and $\vec{v}_{\text{db}}^i$ for the $i$-th keypoint pair $(p_q^i, p_{\text{db}}^i) \in \mathcal{S}_{\text{topo}}$, alongside the relative rotation $\phi_i$, as:

\textcolor{black}{In addition}, we enforce global consistency in scale and orientation on the surviving set $\mathcal{S}_{\text{topo}}$ to \textcolor{black}{further remove structurally inconsistent matches}. Let $\bar{p}_q$ and $\bar{p}_{\text{db}}$ denote the geometric centroids of the keypoints in $\mathcal{S}_{\text{topo}}$ for the query and database modalities, respectively. \textcolor{black}{For} the $i$-th keypoint pair $(p_q^i, p_{\text{db}}^i) \in \mathcal{S}_{\text{topo}}$, we define the heading vectors $\vec{v}_q^i$ and $\vec{v}_{\text{db}}^i$, \textcolor{black}{as well as} the relative rotation $\phi_i$, as \textcolor{black}{follows}:

\begin{equation}
\begin{aligned}
    \vec{v}_q^i &= p_q^i - \bar{p}_q, \quad \vec{v}_{\text{db}}^i = p_{\text{db}}^i - \bar{p}_{\text{db}}, \\
    \phi_i &= \operatorname{ang}(\vec{v}_{\text{db}}^i) - \operatorname{ang}(\vec{v}_q^i),
\end{aligned}
\end{equation}
where $\operatorname{ang}(\cdot)$ denotes the polar angle of a \textcolor{black}{2D vector}. A match is \textcolor{black}{retained only when} it satisfies the global \textcolor{black}{rotation and scale consistency constraints}. The final \textcolor{black}{refined} matching set $\mathcal{S}_{\text{final}}$ is then \textcolor{black}{obtained} as:

\begin{equation}
    \begin{aligned}
    \mathcal{S}_{\text{final}} = \Bigg\{ (p_q^i, p_{\text{db}}^i) \in \mathcal{S}_{\text{topo}} \ \Bigg| \ 
    & \left| \phi_i - \operatorname{med}(\boldsymbol{\phi}) \right| < \epsilon_{\text{ang}} \\
    & \land \left| \frac{\|\vec{v}_{\text{db}}^i\|}{\bar{s} \cdot \|\vec{v}_q^i\|} - 1 \right| \le \epsilon_{\text{scale}} \Bigg\},
    \end{aligned}
\end{equation}
where $\boldsymbol{\phi}$ is the set of relative rotations, $\bar{s}$ is the median scale, and $\epsilon_{\text{ang}}$ and $\epsilon_{\text{scale}}$ are the corresponding tolerance thresholds.

By coupling local topological invariance with global \textcolor{black}{geometric consistency}, this final filtering layer \textcolor{black}{effectively removes} spatially misplaced outliers caused by \textcolor{black}{visually similar structures}.

\textcolor{black}{In summary, C-SATSF progressively refines raw dense correspondences through spatial equalization, texture verification, and structure-consistent refinement, producing a reliable inlier set $\mathcal{S}_{\text{final}}$ for subsequent pose optimization.}

The complete workflow of the C-SATSF mechanism is shown in Algorithm~\ref{alg:CSATSF}.

\subsection{Consensus-Driven Reliability-Aware Position Selection Strategy}

To identify the optimal geographic position from the Top-$N$ candidates and suppress visual decoys, we propose the CD-RAPS strategy. In large-scale search spaces, \textcolor{black}{repetitive global structures may produce deceptively high PnP inlier counts even for incorrect candidate poses}~\cite{lowry2015visual}. To \textcolor{black}{address this issue}, CD-RAPS refines candidate poses through physically constrained optimization, evaluates their multi-dimensional reliability, and \textcolor{black}{selects the final position via} geographic consensus voting.

% For a given candidate $k$, building upon the highly dependable 2D-2D correspondence set $\mathcal{S}_{\text{final}}^{(k)}$ preserved by the preceding filtering module, let $(p_q^i, p_{\text{db}}^i) \in \mathcal{S}_{\text{final}}^{(k)}$ denote the $i$-th matched keypoint pair. By extracting the elevation prior from a DSM at the satellite 2D location $p_{\text{db}}^i$, we lift it to the 3D geographic coordinate $\mathbf{P}_i$. This forms the 2D-3D correspondence set $\mathcal{P} = \{(p_q^i, \mathbf{P}_i)\}$, from which an initial PnP pose is computed~\cite{ye2025exploring}. Instead of directly utilizing this initial solution, we formulate a non-linear optimization guided by physical UAV flight priors~\cite{qin2018vins}. For notational clarity, we omit the candidate index $k$ in the subsequent single-pose derivation.

For a given candidate $k$, we \textcolor{black}{start from the reliable} 2D-2D correspondence set $\mathcal{S}_{\text{final}}^{(k)}$ \textcolor{black}{produced by} the preceding filtering module. Let $(p_q^i, p_{\text{db}}^i) \in \mathcal{S}_{\text{final}}^{(k)}$ denote the $i$-th matched keypoint pair. By \textcolor{black}{querying the DSM at} the satellite 2D location $p_{\text{db}}^i$, we obtain its elevation prior and lift it to the 3D geographic coordinate $\mathbf{P}_i$. This \textcolor{black}{yields} the 2D-3D correspondence set $\mathcal{P} = \{(p_q^i, \mathbf{P}_i)\}$, from which an initial PnP pose is computed~\cite{ye2025exploring}. \textcolor{black}{Rather than directly using} this initial solution, we further formulate a non-linear optimization guided by physical UAV flight priors~\cite{qin2018vins}. For notational clarity, \textcolor{black}{the candidate index $k$ is omitted} in the following single-pose derivation.

% Considering that UAVs in orthographic observation tasks typically employ three-axis stabilized gimbals, the camera's roll angle is physically constrained near zero. We incorporate this hardware-induced limitation, alongside the flight telemetry pitch prior, directly into the optimization objective. This explicitly prevents the solver from converging to physically impossible configurations, thereby guaranteeing the geometric correctness of the estimated pose.

Considering that UAVs in \textcolor{black}{nadir-view observation tasks are often equipped with} three-axis stabilized gimbals, the camera roll angle is \textcolor{black}{usually constrained around zero}. We incorporate this \textcolor{black}{hardware-related constraint}, together with the telemetry pitch prior, into the optimization objective. These physical priors help prevent the solver from \textcolor{black}{compensating erroneous visual matches with unrealistic attitude variations}, thereby improving the \textcolor{black}{stability and plausibility} of the estimated pose.

To implement these physical constraints mathematically, let the camera state be parameterized by $\boldsymbol{\xi} \in \mathfrak{se}(3)$, corresponding to the Lie algebra of the Special Euclidean group $SE(3)$~\cite{sola2018micro}. This serves as a minimal and singularity-free parameterization for continuous optimization, where the corresponding rotation matrix $\mathbf{R}_{\text{wc}}(\boldsymbol{\xi})$ and translation vector $\mathbf{t}_{\text{wc}}(\boldsymbol{\xi})$ are iteratively recovered via the exponential map. Specifically, the optimal pose parameter $\boldsymbol{\xi}$ is obtained by minimizing the joint objective function $E(\boldsymbol{\xi})$:

\begin{equation}
    \begin{aligned}
    E(\boldsymbol{\xi}) =& \sum_{(p_q^i, \mathbf{P}_i) \in \mathcal{P}_{\text{in}}} \left\| p_q^i - \pi(\mathbf{P}_i, \boldsymbol{\xi}) \right\|^2 + \lambda_{\text{roll}} \left( (\mathbf{R}_{\text{wc}}(\boldsymbol{\xi})\mathbf{e}_1)^\top \mathbf{n}_z \right)^2 \\
    &+ \lambda_{\text{pitch}} \left( \theta_{\text{pitch}}(\boldsymbol{\xi}) - \theta_{\text{prior}} \right)^2,
    \end{aligned}
\end{equation}
where $\mathcal{P}_{\text{in}} \subseteq \mathcal{P}$ denotes the robust inlier subset identified during the initial PnP computation, and $\pi(\cdot)$ represents the camera projection function. For the attitude constraints, $\mathbf{e}_1 = [1, 0, 0]^\top$ and $\mathbf{n}_z = [0, 0, 1]^\top$ refer to the camera's lateral axis and the world gravity vector, respectively. The term $\theta_{\text{pitch}}(\boldsymbol{\xi})$ represents the pitch angle analytically derived from the current pose state $\boldsymbol{\xi}$ at each optimization iteration, $\theta_{\text{prior}}$ is the constant pitch angle prior acquired from the UAV's flight telemetry, and $\lambda_{\text{roll}}$ and $\lambda_{\text{pitch}}$ act as \textcolor{black}{fixed attitude-prior weighting coefficients.}

Furthermore, \textcolor{black}{after the optimization converges, we estimate} the geometric positional uncertainty $U_{\text{unc}}$ \textcolor{black}{from} the local curvature of the objective landscape. Let $\mathbf{J}$ denote the Jacobian matrix evaluated at the \textcolor{black}{optimized pose}. The Fisher Information Matrix is approximated as $\mathbf{H} \approx \mathbf{J}^\top \mathbf{J}$. \textcolor{black}{Based on} the Cram\'{e}r-Rao Lower Bound~\cite{sengijpta1995fundamentals}, the posterior covariance matrix $\boldsymbol{\Sigma}_{\xi}$ of the pose parameters and the uncertainty metric $U_{\text{unc}}$ are \textcolor{black}{computed as}:

\begin{equation}
\boldsymbol{\Sigma}_{\xi} = \hat{\sigma}^2 \mathbf{H}^{-1}, \quad U_{\text{unc}} = \sqrt{\text{Tr}(\boldsymbol{\Sigma}_{\text{pos}})},
\end{equation}
where $\hat{\sigma}^2$ denotes the posterior residual variance, $\boldsymbol{\Sigma}_{\text{pos}}$ represents the $3 \times 3$ translational sub-matrix extracted from the full covariance $\boldsymbol{\Sigma}_{\xi}$, and $\operatorname{Tr}(\cdot)$ \textcolor{black}{denotes} the matrix trace operator. To ensure dimensional consistency, $U_{\text{unc}}$ \textcolor{black}{summarizes the multidimensional pose uncertainty as} a scalar metric representing the positional standard deviation. Physically, a high $U_{\text{unc}}$ \textcolor{black}{indicates} a flat objective landscape and an unstable solution, e.g., caused by thermal image blur or low contrast, whereas a low value \textcolor{black}{corresponds to} a sharp, well-constrained minimum.

To avoid unreliable pose hypotheses, a hard gating mechanism rejects candidates that fail to yield a convergent solution, contain insufficient inliers, or produce singular Hessian matrices with infinite uncertainty~\cite{zhang2016degeneracy}.

% For the surviving valid candidates, we holistically evaluate their reliability across four complementary dimensions: the global cosine similarity $A_{\text{ret}}$ inherited from the initial semantic retrieval phase, the geometric inlier count $N_{\text{in}}$ preserved post-optimization, the final reprojection error $E_{\text{err}}$, and the derived geometric uncertainty $U_{\text{unc}}$. To ensure comparable scales, we apply a robust Min-Max normalization to map each metric into a unified confidence scalar ranging from zero to one. Positive indicators ($A_{\text{ret}}$ and $N_{\text{in}}$) are positively normalized, whereas negative indicators ($E_{\text{err}}$ and $U_{\text{unc}}$) are inverted to unify the evaluation criteria, ensuring that higher values consistently indicate greater reliability. Ultimately, a base reliability metric $R_{\text{base}}^{(k)}$ for the $k$-th candidate is aggregated via a weighted fusion:
For the surviving candidates, we evaluate their reliability across four dimensions: the semantic cosine similarity $A_{\text{ret}}$, the post-optimization inlier count $N_{\text{in}}$, the reprojection error $E_{\text{err}}$, and the geometric uncertainty $U_{\text{unc}}$. To ensure scale consistency, these metrics are Min-Max normalized to $[0, 1]$. By inverting the error-based metrics ($E_{\text{err}}$ and $U_{\text{unc}}$), we align all indicators such that higher values denote greater reliability. Ultimately, the base reliability $R_{\text{base}}^{(k)}$ for the $k$-th candidate is aggregated via weighted fusion:

\begin{equation}
R_{\text{base}}^{(k)} = w_1 A_{\text{ret}}^{(k)} + w_2 N_{\text{in}}^{(k)} + w_3 E_{\text{err}}^{(k)} + w_4 U_{\text{unc}}^{(k)},
\end{equation}
% where $\{w_1, w_2, w_3, w_4\}$ are non-negative weighting coefficients that govern the relative contribution of each reliability dimension, subject to the normalization constraint $\sum_{i=1}^{4} w_i = 1$. These trade-off parameters are empirically configured to balance the influence among semantic retrieval fidelity, geometric inlier abundance, spatial reprojection accuracy, and pose estimation stability.
where $\{w_i\}_{i=1}^4$ are empirically determined non-negative weights satisfying $\sum_{i=1}^{4} w_i = 1$.

\begin{algorithm}[!t]
\DontPrintSemicolon
\caption{Consensus-Driven Reliability-Aware Position Selection Strategy}
\label{alg:CDRAPS}

% 定义带有左侧竖线包裹的自定义块
\SetKwBlock{PartThree}{\textbf{Part 3: Optimal Selection}}{}

\SetKwInOut{Input}{Input}
\SetKwInOut{Output}{Output}

\Input{
  Top-$N$ inlier sets $\{\mathcal{S}_{\text{final}}^{(k)}\}_{k=1}^N$ and retrieval scores $\{A_{\text{ret}}^{(k)}\}_{k=1}^N$; Aligned DSM Database $\mathcal{D}_{\text{DSM}}$.
}
\Output{
  Optimal Absolute Horizontal Position $(x^*, y^*)$.
}

\BlankLine
\textbf{Part 1: Physically Constrained Pose Optimization}\;
\For{$k = 1$ \KwTo $N$}{
    $\mathcal{P} \leftarrow \text{LiftTo3D}(\mathcal{S}_{\text{final}}^{(k)}, \mathcal{D}_{\text{DSM}})$\;
    $\boldsymbol{\xi}_{\text{init}}, \mathcal{P}_{\text{in}} \leftarrow \text{ComputeInitialPnP}(\mathcal{P})$\;
    $N_{\text{in}}^{(k)} \leftarrow |\mathcal{P}_{\text{in}}|$\;
    $\boldsymbol{\xi}_k \leftarrow \arg\min_{\boldsymbol{\xi}} E(\boldsymbol{\xi} \mid \boldsymbol{\xi}_{\text{init}})$\;
    $E_{\text{err}}^{(k)} \leftarrow E(\boldsymbol{\xi}_k)$\;
    $U_{\text{unc}}^{(k)} \leftarrow \sqrt{\text{Tr}\left( \big[ \hat{\sigma}^2 \mathbf{H}(\boldsymbol{\xi}_k)^{-1} \big]_{\text{pos}} \right)}$\;
    \vspace{4pt}
    
    $\mathbf{L}_k \leftarrow \mathbf{t}_{\text{wc}}(\boldsymbol{\xi}_k)$\;
    $R_{\text{base}}^{(k)} \leftarrow \text{BaseReliability}(A_{\text{ret}}^{(k)}, N_{\text{in}}^{(k)}, E_{\text{err}}^{(k)}, U_{\text{unc}}^{(k)})$\;
}

\BlankLine
\textbf{Part 2: Geographic Consensus Voting}\;
\For{$k = 1$ \KwTo $N$}{
    $C_{\text{geo}}^{(k)} \leftarrow \sum_{j \in \mathcal{N}_k \mid R_{\text{base}}^{(j)} \ge \tau} R_{\text{base}}^{(j)} \cdot \left( 1 - \frac{\|\mathbf{L}_k - \mathbf{L}_j\|_2}{d_{\text{max}}} \right)$\;
    \vspace{4pt}
    $R_{\text{total}}^{(k)} \leftarrow R_{\text{base}}^{(k)} + \min(\omega_{\text{geo}}C_{\text{geo}}^{(k)}, \omega_{\text{base}}R_{\text{base}}^{(k)})$\;
}

\BlankLine
\PartThree{
    $k^* \leftarrow \arg\max_k R_{\text{total}}^{(k)}$\;
    $(x^*, y^*) \leftarrow \mathbf{L}_{k^*}$\;
}

\Return{$(x^*, y^*)$}\;

\end{algorithm}

% Finally, to exploit the geographic consensus that correct candidates cluster tightly while incorrect visual decoys scatter randomly, we introduce a distance-decaying spatial consistency voting mechanism. For a valid candidate $k$ located at geographic coordinate $\mathbf{L}_k$, its spatial neighborhood $\mathcal{N}_k$ comprises the nearest candidates within a specific geographic distance threshold $d_{\text{max}}$. Candidates within this radius contribute a weighted geographic consensus score $C_{\text{geo}}^{(k)}$ to candidate $k$, penalizing distant neighbors linearly:

Finally, to exploit the geographic consensus that correct candidates \textcolor{black}{tend to form compact spatial clusters, whereas incorrect visual decoys are usually more dispersed}, we introduce a distance-decaying spatial consistency voting mechanism. For a valid candidate $k$ located at geographic coordinate $\mathbf{L}_k$, its spatial neighborhood $\mathcal{N}_k$ \textcolor{black}{is defined as the set of nearby candidates} within a geographic distance threshold $d_{\text{max}}$. Candidates within this radius contribute to the geographic consensus score $C_{\text{geo}}^{(k)}$ of candidate $k$, \textcolor{black}{with closer neighbors assigned higher weights}:

\begin{equation}
    C_{\text{geo}}^{(k)} = \sum_{j \in \mathcal{N}_k, R_{\text{base}}^{(j)} \ge \tau} R_{\text{base}}^{(j)} \cdot \left( 1 - \frac{\|\mathbf{L}_k - \mathbf{L}_j\|_2}{d_{\text{max}}} \right),
\end{equation}
where $\tau$ denotes the minimum reliability threshold that exclusively accepts contributions from high-confidence neighbors, effectively preventing score accumulation from low-quality outliers. Subsequently, the total reliability score $R_{\text{total}}^{(k)}$ for each candidate is formulated by fusing the base reliability with the geographic consensus reward:

\begin{equation}
    R_{\text{total}}^{(k)} = R_{\text{base}}^{(k)} + \min\left(\omega_{\text{geo}} C_{\text{geo}}^{(k)}, \omega_{\text{base}} R_{\text{base}}^{(k)}\right),
\end{equation}
where $\omega_{\text{geo}}$ and $\omega_{\text{base}}$ are scaling factors controlling the consensus reward. Ultimately, the final position index $k^*$ is determined by selecting the candidate with the highest total reliability score. The final UAV horizontal position $(x^*, y^*)$ is then explicitly retrieved from the pre-computed spatial location \textcolor{black}{$\mathbf{L}_{k^*}$} of this selected optimal candidate.

Through this synergy of physically constrained optimization, multi-dimensional reliability evaluation, and geographic consensus, the proposed CD-RAPS strategy \textcolor{black}{effectively suppresses visual decoys}, thereby completing a robust UAV geo-localization pipeline. 

The complete workflow of the CD-RAPS strategy is shown in Algorithm~\ref{alg:CDRAPS}.

\begin{figure*}[!t]
    \centering
    \includegraphics[width=\linewidth]{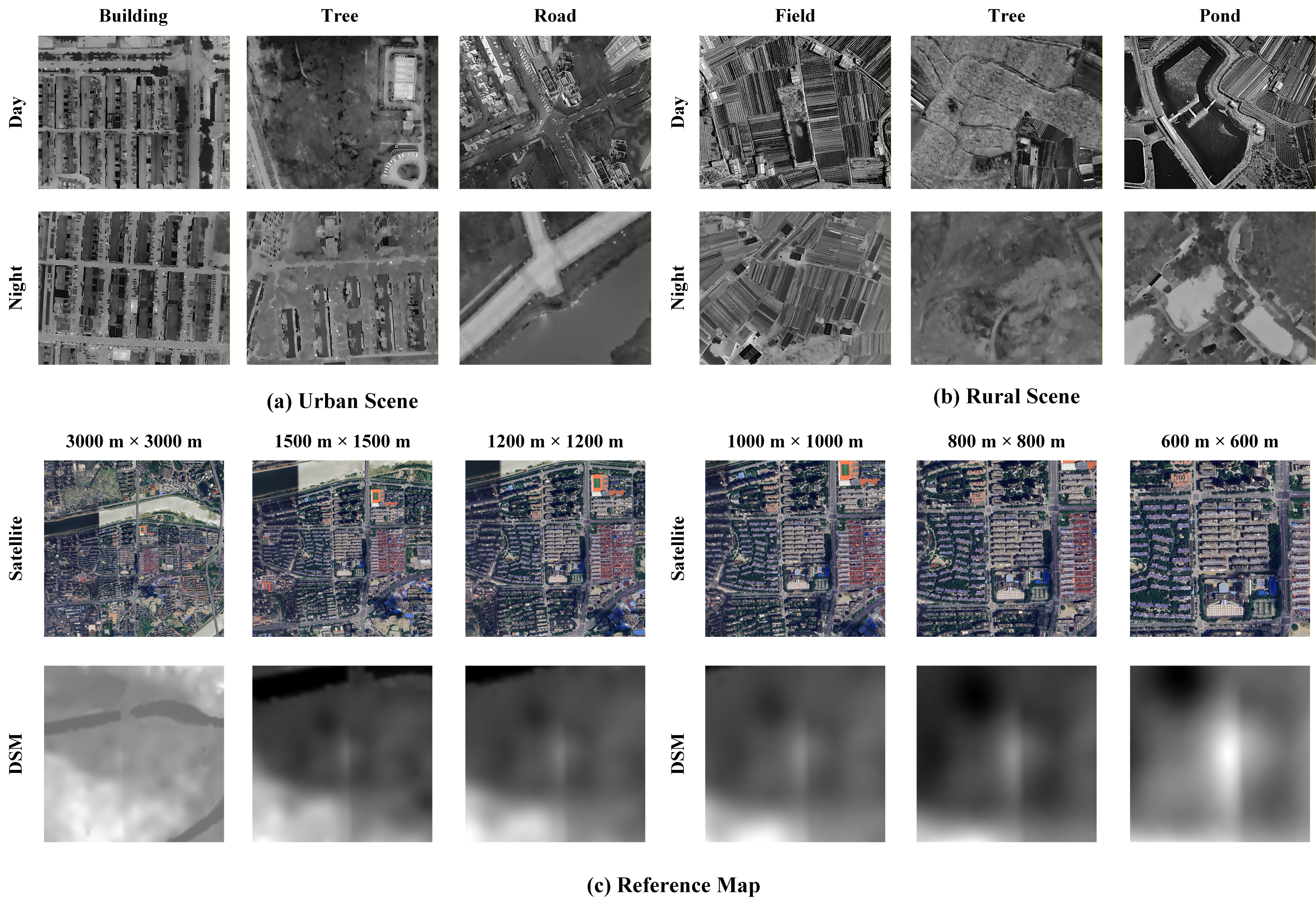}
    \caption{Visual samples from the constructed Thermal-UAV dataset. The dataset systematically captures profound modality discrepancies and diurnal thermal variations across diverse spatial topologies: (a) Urban and (b) Rural Scenes showcase thermal UAV queries of distinct semantic categories (e.g., buildings, roads, fields) during daytime and nighttime; (c) Reference Map presents the global database at varying search scales, comprising visible-light satellite ortho-photos and their spatially aligned Digital Surface Models (DSM) to supply crucial 3D elevation priors.}
    \label{Dataset}
\end{figure*}

\section{Experimental Setup} \label{Experimental Setup}

\subsection{Datasets}

\textcolor{black}{
To address the scarcity of large-scale benchmarks for cross-modal UAV geo-localization, we construct the Thermal-UAV dataset, which contains 11,890 thermal images collected by a DJI Matrice 4T drone\footnote{\url{https://www.dji.com/support/product/}} at altitudes of 300--350~m under a nadir-view configuration. The dataset was collected in Changsha, China, where the satellite and DSM reference maps cover a city-scale area of approximately 98.74~km$^2$. Within this reference area, UAV thermal trajectories are distributed across representative urban and rural subregions, including dense buildings, roads, agricultural fields, and vegetation. The sequences were mainly acquired from December to March, corresponding to winter and early-spring conditions, and include both daytime and nighttime flights to capture diurnal thermal variations.
}

\textcolor{black}{
As shown in Fig.~\ref{Dataset}(a) and (b), the representative thermal queries reveal substantial thermal-visible modality gaps under different environmental conditions. The dataset is organized into dense multi-segment flight trajectories and partitioned into training (8,115), validation (1,425), and testing (2,350) sets. The global reference database is constructed as a single continuous map, consisting of Google Earth satellite imagery\footnote{\url{https://www.google.com/maps}} at 0.26~m/pixel and a spatially aligned DSM at 5.29~m/pixel. All satellite ortho-images and DSM tiles are geo-referenced under the same coordinate system to support metric-level localization and 2D-3D pose estimation. Fig.~\ref{Dataset}(c) visualizes the unified reference map and retrieval scenarios under different search-space sizes.
}

\textcolor{black}{
We do not evaluate on the existing Boson-nighttime dataset~\cite{xiao2023long}, because it lacks spatially aligned DSM data, which is required for absolute pose estimation via 2D-3D PnP. Therefore, Thermal-UAV provides a necessary benchmark for evaluating 3D-aware cross-modal UAV geo-localization pipelines.
}

\subsection{Evaluation Metrics}
We evaluate SCC-Loc across two stages: coarse retrieval and fine-grained localization. For coarse retrieval, we use \textbf{Recall@$\mathbf{N}$} ($N \in \{3, 5, 10\}$)~\cite{xiao2024sthn}. A retrieval is deemed a hit if the Position Deviation Error (PDE)—the normalized Euclidean distance between the retrieved patch center and the ground truth—is strictly less than 0.5~\cite{ye2025exploring}. 

For fine-grained absolute position estimation, we measure geometric precision using \textbf{Acc@$\mathbf{R}$} ($R \in \{5, 10, 20\}$) and \textbf{ME $\pm$ SD}~\cite{dhaouadi2025ortholoc}. \textbf{Acc@$\mathbf{R}$} represents the percentage of predicted coordinates falling within $R$ meters of the ground truth. Additionally, we report the Mean Error (ME) and Standard Deviation (SD) in meters to quantify the overall metric accuracy and spatial robustness, respectively. 

Finally, to assess deployment feasibility, we report the average \textbf{Inference Time} (seconds/query) and peak GPU \textbf{Memory Consumption} (MB)~\cite{li2025geovins}. Specifically, the memory metric accounts for the model's parameter footprint and inference overhead, excluding the reference database and data loading memory.

\subsection{Baseline Methods}
To evaluate SCC-Loc against the thermal-visible modality gap, we construct comprehensive baselines across three categories. First, for \textbf{coarse retrieval}, we employ CAMP~\cite{wu2024camp} (visible-only trained) to assess zero-shot degradation, alongside foundation models DINOv2~\cite{oquab2023dinov2} and DINOv3~\cite{simeoni2025dinov3} to test their inherent modality-invariant semantics. Second, for \textbf{fine-grained matching}, we couple these retrievers with varying feature matchers: RoMa~\cite{edstedt2024roma} (an intra-modal dense matcher) to establish a baseline for performance degradation under severe spectral discrepancies; XoFTR~\cite{tuzcuouglu2024xoftr}, a fast semi-dense matcher \textcolor{black}{designed for cross-modal matching}; and MINIMA$_{\text{RoMa}}$~\cite{ren2025minima}, which explicitly fine-tunes RoMa on a cross-modal benchmark to serve as a domain-adapted upper bound. Finally, we benchmark against \textbf{domain-specific networks} STHN~\cite{xiao2024sthn} and NIVnet~\cite{liu2024multimodal}. For fairness, both are retrained on our Thermal-UAV training split. Despite relying on planar homography regression rather than our zero-shot 3D-aware formulation, \textcolor{black}{these methods serve as strong supervised baselines for thermal-to-visible localization}.

\subsection{Implementation Details}

\textbf{Implementation Platform.} We implement SCC-Loc in PyTorch, executing all experiments on a workstation equipped with an Intel Core i5-14600K CPU and an NVIDIA RTX 4060 Ti GPU.

\textcolor{black}{\textbf{Geo-Spatial Preprocessing.} Assuming that UAV altitude, pitch, and yaw are available as onboard priors, we use the yaw prior to rotate the satellite map into approximate alignment with the UAV thermal query, thereby reducing large in-plane rotation differences. To maintain scale consistency, each satellite crop is determined according to the physical ground coverage and Ground Sampling Distance (GSD) of the UAV query~\cite{ye2025exploring}. Its crop size is further enlarged by a scaling factor $s_{\text{gsd}}$ to preserve more peripheral context for subsequent fine-grained matching.}

\textbf{Hyperparameter Settings.} For coarse retrieval, the continuous satellite ortho-photo is discretized into a searchable patch database using a sliding window with an overlap ratio $\rho_{\text{overlap}}$ within a geographic search area $A_{\text{search}}$. To ensure full reproducibility, all hyperparameters governing the entire pipeline are detailed in Table~\ref{tab:hyperparameters}. Unless otherwise specified, these default settings remain strictly constant across all evaluations.

% \textbf{Baseline Training Protocol.}
% \textcolor{black}{
% Since SCC-Loc is a training-free framework, it does not use the training or validation split for model optimization, domain-specific fine-tuning, or parameter learning. In contrast, the supervised domain-specific baselines, including STHN~\cite{xiao2024sthn} and NIVnet~\cite{liu2025nivnet}, are trained from scratch on the Thermal-UAV training set for fair comparison. Specifically, the 8,115 training images are used to optimize the model parameters, the 1,425 validation images are used for hyperparameter tuning and checkpoint selection, and the 2,350 testing images are kept strictly unseen for final evaluation. Both baselines are trained using paired thermal-satellite samples with valid ground-truth geometric displacement labels, and invalid pairs are filtered during training and evaluation. The checkpoint with the best validation localization performance is selected for final testing. All learning-based baselines follow the same Thermal-UAV data split and are evaluated under the same search-area settings as SCC-Loc.
% }

% \textcolor{black}{
% Both baselines are optimized with the Adam optimizer using the same input resolution and batch size setting. During training, we monitor the geometric regression loss and the validation localization metrics, including mean localization error and Recall@5/10/20. The final reported results are obtained from the checkpoint that achieves the best validation performance.
% }

\begin{figure*}[!t]
    \centering
    \includegraphics[width=\linewidth]{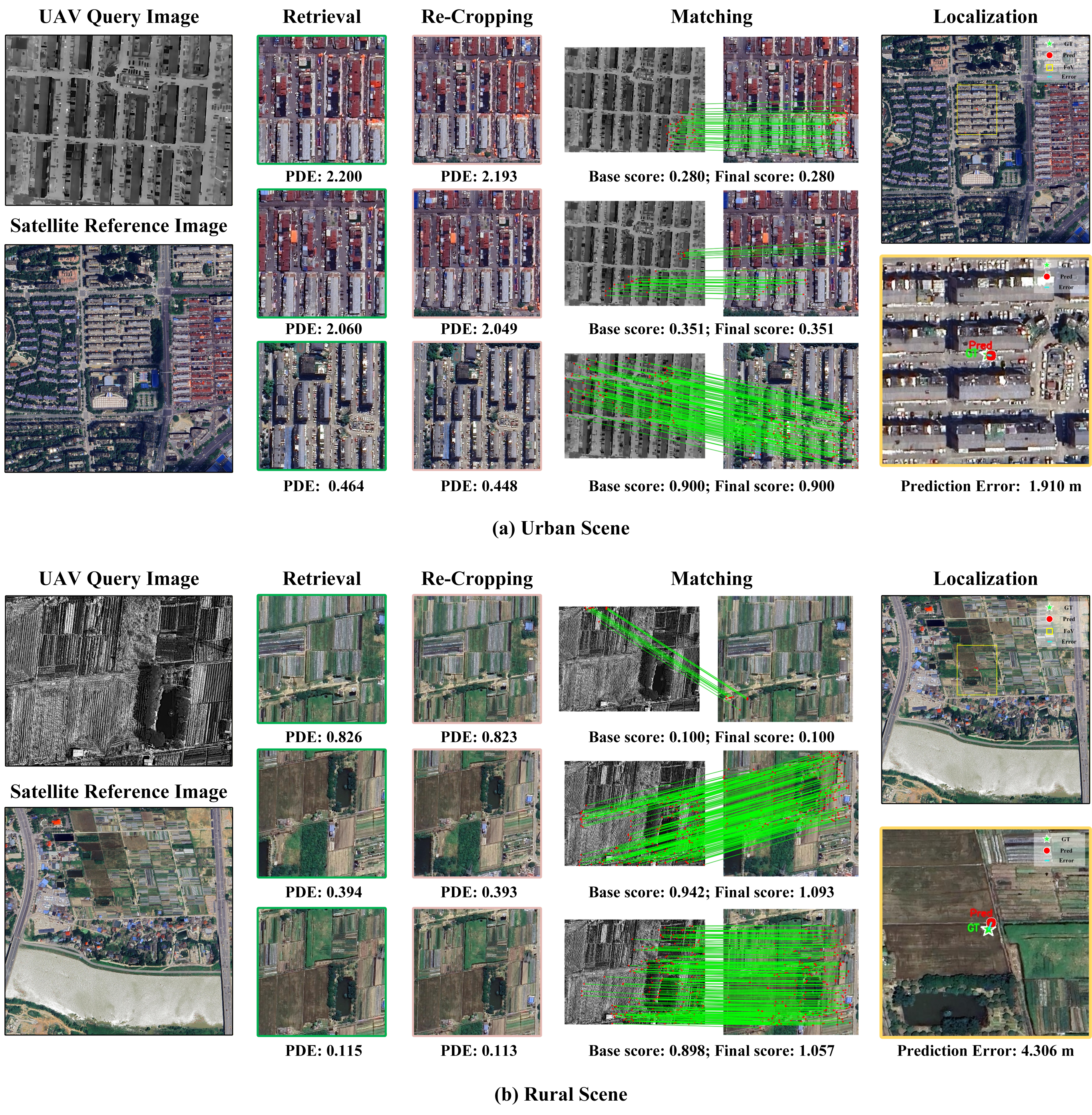}
    \caption{Qualitative visualization of the proposed SCC-Loc pipeline in (a) Urban and (b) Rural scenarios. The process illustrates the adaptive correction of spatial quantization bias via the SGVA module (\textbf{Re-Cropping}) and the progressive suppression of structural outliers using the C-SATSF mechanism (\textbf{Matching}). Finally, based on the purified correspondences, the CD-RAPS strategy refines candidate poses via physically constrained optimization, and computes the total reliability (\textbf{Final score}) by fusing the multi-dimensional evaluation (\textbf{Base score}) with geographic consensus to determine the robust optimal hypothesis for precise \textbf{Localization}.}
    \label{Quantitative_vis}
\end{figure*}

\textcolor{black}{
\textbf{Baseline Training Protocol.}
For all training-free coarse-to-fine methods and the proposed SCC-Loc, no training images are used for model optimization, domain-specific fine-tuning, or parameter learning. These methods are evaluated in a zero-shot manner by directly using their pre-trained retrieval and matching models, with all hyperparameters fixed before testing. The 2,350 testing images are used only for final evaluation.}

\textcolor{black}{
For supervised domain-specific baselines, including STHN and NIVnet, we train all models from scratch on Thermal-UAV using the same training/validation/testing split. Specifically, the 8,115 training images are used for parameter optimization, the 1,425 validation images are used for hyperparameter tuning and checkpoint selection, and the 2,350 testing images are kept strictly unseen for final evaluation. STHN is trained with AdamW, OneCycleLR scheduling, a maximum learning rate of $3\times10^{-5}$, weight decay of $1\times10^{-5}$, batch size 16, and $256\times256$ input resolution. NIVnet is trained with AdamW, warmup-cosine scheduling, an initial learning rate of $1\times10^{-4}$, weight decay of $1\times10^{-5}$, and batch size 4. For both supervised baselines, the checkpoint with the best validation localization performance is selected for final testing.
}

\begin{table}[!t] % 放在单栏顶部
\centering
\caption{Detailed hyperparameter configurations for the proposed SCC-Loc framework.}
\label{tab:hyperparameters}

% 调节行高与列距
\renewcommand{\arraystretch}{1.2} 
\setlength{\tabcolsep}{8pt} 

% 核心魔法：自适应单栏宽度
\resizebox{\columnwidth}{!}{%
\begin{tabular}{@{}lcc@{}}
\toprule
\textbf{Hyperparameter Description} & \textbf{Symbol} & \textbf{Value} \\
\midrule

% --- Retrieval 模块子表头 ---
\multicolumn{3}{@{}l}{\textbf{\textit{Coarse Retrieval Phase}}} \\
\quad Geographic search area & $A_{\text{search}}$ & $600 \times 600\text{ m}^2$ \\
\quad Sliding window overlap & $\rho_{\text{overlap}}$ & $60\%$ \\
\quad GSD scaling factor & $s_{\text{gsd}}$ & $1.5$ \\
\quad Minimum error tolerance & $\epsilon_{\text{min}}$ & $10^{-6}$ \\
\quad Pooling exponent & $\psi$ & $4$ \\
\addlinespace % 留出透明的呼吸缝隙

% --- SGVA 模块子表头 ---
% \multicolumn{3}{@{}l}{\textbf{\textit{Semantic-Guided Viewport Alignment (SGVA)}}} \\
% \quad Sensitivity factor & $\lambda$ & $5.0$ \\
% \quad Confidence boost & $\alpha$ & $0.5$ \\
% \quad Expansion rate & $\beta$ & $0.2$ \\
% \addlinespace

\multicolumn{3}{@{}l}{\textcolor{black}{\textbf{\textit{Semantic-Guided Viewport Alignment (SGVA)}}}} \\
\quad \textcolor{black}{Expansion rate} & \textcolor{black}{$\alpha$} & \textcolor{black}{$0.2$} \\
\quad \textcolor{black}{Center-shift strength} & \textcolor{black}{$\beta$} & \textcolor{black}{$1.0$} \\
\addlinespace

% --- C-SATSF 模块子表头 ---
\multicolumn{3}{@{}l}{\textbf{\textit{Cascaded Filtering Mechanism (C-SATSF)}}} \\
\quad Grid partition size & $G$ & $8 \times 8$ \\
\quad Min. reservation limit & $Q_{\text{base}}$ & $3$ \\
\quad Saliency threshold scalar & $\gamma$ & $0.5$ \\
\quad Topological tolerance & $\epsilon_{\text{topo}}$ & $0.4$ \\
\quad Angular tolerance & $\epsilon_{\text{ang}}$ & $20^{\circ}$ \\
\quad Scale tolerance & $\epsilon_{\text{scale}}$ & $0.3$ \\
\addlinespace

% --- CD-RAPS 模块子表头 ---
\multicolumn{3}{@{}l}{\textbf{\textit{Consensus-Driven Selection (CD-RAPS)}}} \\
\quad Roll constraint weight & $\lambda_{\text{roll}}$ & $1000$ \\
\quad Pitch constraint weight & $\lambda_{\text{pitch}}$ & $15$ \\
\quad Reliability weights & $\{w_i\}_{i=1}^{4}$ & $\{0.1, 0.2, 0.35, 0.35\}$ \\
\quad Max. geographic distance & $d_{\text{max}}$ & $20\text{ m}$ \\
\quad Min. reliability threshold & $\tau$ & $0.3$ \\
\quad Geo-consensus weight & $\omega_{\text{geo}}$ & $0.2$ \\
\quad Base reliability weight & $\omega_{\text{base}}$ & $0.5$ \\

\bottomrule
\end{tabular}%
}
\end{table}

\begin{table*}[!t]
\centering
\caption{Quantitative comparison of the proposed SCC-Loc framework against various two-stage baselines, which comprise different combinations of retrieval and matching models. The best results are highlighted in \textbf{bold}.}
\label{tab:localization_results}

% ==================== ��️ 全局行高与列距调节器 ====================
\renewcommand{\arraystretch}{1.25} % 回调到1.25，避免1.5带来的松散感
\setlength{\tabcolsep}{2pt}        % 放宽列距，增加数字和符号之间的呼吸感
% ================================================================

\resizebox{\textwidth}{!}{%
\begin{tabular}{l cccc cccc cccc}
\toprule
\multirow{2}{*}{\textbf{Method}} & \multicolumn{4}{c}{\textbf{Top-3 Candidates}} & \multicolumn{4}{c}{\textbf{Top-5 Candidates}} & \multicolumn{4}{c}{\textbf{Top-10 Candidates}} \\
\cmidrule(lr){2-5} \cmidrule(lr){6-9} \cmidrule(lr){10-13}
& \textbf{Recall@3} & \textbf{Acc@5/10/20} & \textbf{ME $\pm$ SD} & \textbf{Time} 
& \textbf{Recall@5} & \textbf{Acc@5/10/20} & \textbf{ME $\pm$ SD} & \textbf{Time} 
& \textbf{Recall@10} & \textbf{Acc@5/10/20} & \textbf{ME $\pm$ SD} & \textbf{Time} \\
\midrule

% ---------- 第一组：CAMP 阵营基线 ----------
CAMP + XoFTR
& 75.15 & 1.53 / 5.96 / 16.51 & $280.52 \pm 374.93$ & 1.03
& 85.70 & 1.66 / 6.94 / 20.04 & $193.75 \pm 306.82$ & \textbf{1.24}
& 96.13 & 1.66 / 8.13 / 23.79 & $121.00 \pm 216.86$ & \textbf{1.79} \\

CAMP + RoMa
& 75.15 & 3.06 / 12.04 / 31.49 & $91.89 \pm 139.31$ & 4.01
& 85.70 & 4.26 / 14.00 / 36.98 & $69.00 \pm 105.93$ & 6.30
& 96.13 & 4.38 / 15.83 / 39.70 & $49.17 \pm 82.95$ & 11.97 \\

CAMP + MINIMA$_{\text{RoMa}}$   
& 75.15 & 4.72 / 15.57 / 40.43 & $84.26 \pm 131.58$   & 4.22
& 85.70 & 4.77 / 17.62 / 46.17 & $61.79 \pm 111.69$   & 6.60
& 96.13 & 6.77 / 21.11 / 49.83 & $40.92 \pm 78.34$ & 12.53 \\

\midrule % ----- 分界线：显著区分检索模型的不同 -----

% ---------- 第二组：DINOv2 阵营基线 ----------
DINOv2 + XoFTR
& 91.23 & 1.66 / 7.57 / 21.45  & $185.70 \pm 307.10$  & \textbf{0.92}
& 96.77 & 2.17 / 8.64 / 24.51  & $141.61 \pm 254.98$  & 1.30
& 99.45 & 2.30 / 9.57 / 25.87  & $100.86 \pm 188.14$ & 1.96 \\

DINOv2 + RoMa                   
& 91.23 & 3.57 / 14.98 / 39.28 & $55.21 \pm 81.07$    & 4.28
& 96.77 & 4.09 / 16.81 / 43.45 & $48.04 \pm 70.72$    & 6.36
& 99.45 & 4.89 / 16.85 / 42.21 & $40.50 \pm 56.01$ & 12.02 \\

DINOv2 + MINIMA$_{\text{RoMa}}$ 
& 91.23 & 5.62 / 19.83 / 50.04 & $52.17 \pm 95.75$    & 4.21
& 96.77 & 6.09 / 20.51 / 52.47 & $37.82 \pm 67.01$    & 6.43
& 99.45 & 6.85 / 22.85 / 54.09 & $31.32 \pm 51.18$ & 12.28 \\

\midrule 

% ---------- 第三组：DINOv3 阵营基线 ----------
DINOv3 + XoFTR 
& 91.66 & 1.66 / 7.11 / 21.36 & $172.17 \pm 291.38$   & 1.20
& 96.77 & 2.04 / 8.21 / 23.32 & $128.96 \pm 233.40$   & 1.41
& 99.53 & 2.17 / 9.36 / 25.57 & $95.59 \pm 177.26$ & 2.19 \\

DINOv3 + RoMa
& 91.66 & 3.87 / 14.38 / 38.85 & $57.81 \pm 82.84$    & 4.20
& 96.77 & 5.02 / 16.68 / 42.77 & $47.77 \pm 71.82$    & 6.52
& 99.53 & 5.02 / 18.72 / 44.64 & $40.09 \pm 57.15$ & 12.11 \\

DINOv3 + MINIMA$_{\text{RoMa}}$
& 91.66 & 5.49 / 18.68 / 49.15 & $50.82 \pm 91.57$    & 4.34
& 96.77 & 4.89 / 20.21 / 53.06 & $37.66 \pm 67.04$    & 7.41
& 99.53 & 6.64 / 20.81 / 53.45 & $29.19 \pm 43.97$ & 12.53 \\

% \midrule

% % ---------- 第四组：AnyLoc & UAVPlace 阵营基线 ----------
% AnyLoc + MINIMA$_{\text{RoMa}}$
% & 2.14 & 2.14 / 2.14 / 2.14 & $2.14 \pm 2.14$    & 2.14
% & 2.14 & 2.14 / 2.14 / 2.14 & $2.14 \pm 2.14$    & 2.14
% & 2.14 & 2.14 / 2.14 / 2.14 & $2.14 \pm 2.14$ & 2.14 \\

% UAVPlace + MINIMA$_{\text{RoMa}}$
% & 2.14 & 2.14 / 2.14 / 2.14 & $2.14 \pm 2.14$    & 2.14
% & 2.14 & 2.14 / 2.14 / 2.14 & $2.14 \pm 2.14$    & 2.14
% & 2.14 & 2.14 / 2.14 / 2.14 & $2.14 \pm 2.14$ & 2.14 \\

\midrule % ----- 决战分界线：隆重引出本文方法 -----
% ---------- 第五组：本文提出方法 ----------
\textbf{SCC-Loc (Ours)}         
& \textbf{92.30} & \textbf{33.49} / \textbf{56.43} / \textbf{76.64} & $\mathbf{26.28 \pm 67.94}$ & 7.55
& \textbf{96.98} & \textbf{51.66} / \textbf{81.32} / \textbf{89.87} & $\mathbf{15.30 \pm 44.54}$ & 7.52
& \textbf{99.57} & \textbf{52.09} / \textbf{86.38} / \textbf{93.53} & $\mathbf{9.37 \pm 25.26}$ & 14.21 \\
\bottomrule
\end{tabular}%
}
\end{table*}

\begin{table*}[!t]
\centering
\caption{Comparison with domain-specific baselines across varying search areas on our Thermal-UAV dataset under the Top-10 retrieval setting. Best results are in \textbf{bold}.}
\label{tab:separate_comparison}

% ==================== ��️ 顶刊宽表美学调节器 ====================
\renewcommand{\arraystretch}{1.25} % 保持舒适的行高
\setlength{\tabcolsep}{3.5pt}      % 横向展开后，微调列距让表格铺满双栏且不拥挤
% ================================================================

\resizebox{\textwidth}{!}{%
\begin{tabular}{l ccc ccc ccc c}
\toprule
\multirow{2}{*}{\textbf{Method}} 
& \multicolumn{3}{c}{\textbf{Search Area: $600\times600$}} 
& \multicolumn{3}{c}{\textbf{Search Area: $800\times800$}} 
& \multicolumn{3}{c}{\textbf{Search Area: $1200\times1200$}} 
& \multirow{2}{*}{\textbf{Memory}} \\
\cmidrule(lr){2-4} \cmidrule(lr){5-7} \cmidrule(lr){8-10}
& \textbf{Acc@5/10/20} & \textbf{ME $\pm$ SD} & \textbf{Time} 
& \textbf{Acc@5/10/20} & \textbf{ME $\pm$ SD} & \textbf{Time} 
& \textbf{Acc@5/10/20} & \textbf{ME $\pm$ SD} & \textbf{Time} & \\
\midrule

% ---------- 领域特定基线：NIVnet ----------
NIVnet 
& 0.10 / 0.80 / 2.60 & $91.64 \pm 35.06$ & \textbf{0.02}
& 0.10 / 0.60 / 2.10 & $92.37 \pm 35.63$ & \textbf{0.03}
& 0.20 / 0.70 / 1.90 & $99.21 \pm 47.49$ & \textbf{0.04} & 15 \\

% ---------- 领域特定基线：STHN ----------
STHN        
& 0.50 / 3.10 / 11.50 & $40.70 \pm 15.49$ & 0.03
& 0.70 / 3.10 / 12.10 & $38.76 \pm 16.57$ & \textbf{0.03}
& 1.00 / 3.40 / 12.00 & $49.60 \pm 44.40$ & \textbf{0.04} & \textbf{5.5} \\

% ---------- 强基线：DINOv2 + MINIMA_RoMa ----------
DINOv2 + MINIMA$_{\text{RoMa}}$
& 6.85 / 22.85 / 54.09 & $31.32 \pm 51.18$ & 12.28
& 6.26 / 20.89 / 52.51 & $36.92 \pm 73.59$ & 12.50
& 5.11 / 19.06 / 48.13 & $69.05 \pm 148.08$ & 13.62 & 2335.5 \\

\midrule % 粗线压阵，隆重引出本文方法

% ---------- 本文方法：SCC-Loc ----------
\textbf{SCC-Loc (Ours)}
& \textbf{52.09} / \textbf{86.38} / \textbf{93.53} & $\mathbf{9.37 \pm 25.26}$ & 14.21
& \textbf{51.40} / \textbf{81.70} / \textbf{89.87} & $\mathbf{16.84 \pm 55.63}$ & 14.30
& \textbf{46.30} / \textbf{76.00} / \textbf{83.45} & $\mathbf{46.62 \pm 127.04}$ & 16.12 & 1174.7 \\

\bottomrule
\end{tabular}%
}
\end{table*}

\section{Results} \label{Results}

\subsection{Illustrative Operation}

\textcolor{black}{To provide an intuitive understanding of the SCC-Loc pipeline, Fig.~\ref{Quantitative_vis} presents the stage-by-stage localization process in two challenging cross-modal scenarios, including urban and rural scenes. In each scenario, the UAV query is localized within an $800 \times 800$~m satellite reference area, with the GSD scaling factor fixed to 1. We further visualize the evolution of the Top-3 retrieved candidates throughout retrieval, semantic viewport alignment, dense matching, and final localization, thereby illustrating how SCC-Loc progressively corrects spatial misalignment, suppresses unreliable correspondences, and selects the final geographic position.}

{\color{black}
\textbf{1) Adaptive Viewport and Dense Matching:}
At the initial retrieval stage, spatial quantization bias can shift the true UAV-visible region toward the boundary of the retrieved satellite patch, reducing the effective overlap for fine-grained matching. As shown in the \textbf{Re-Cropping} columns, SGVA adaptively realigns the satellite viewport using the semantic response distribution, leading to improved candidate-query alignment and lower PDE, e.g., from 0.464 to 0.448 for the Top-3 urban candidate. During the \textbf{Matching} stage, MINIMA$_{\text{RoMa}}$ first establishes dense cross-modal correspondences, after which C-SATSF filters them by retaining geometrically consistent inliers, shown as green lines, and suppressing unreliable matches induced by repetitive structures and visual decoys.

\textbf{2) Suppressing Decoys via Geographic Consensus:}
The effectiveness of CD-RAPS can be observed by comparing the \textbf{Base score} and the \textbf{Final score} in the \textbf{Localization} stage. Repetitive scene structures, such as similar residential rooftops in Fig.~\ref{Quantitative_vis}(a) and visually analogous agricultural plots in Fig.~\ref{Quantitative_vis}(b), can act as visual decoys. These incorrect candidates may obtain moderate base scores due to local feature similarity and plausible dense correspondences. However, since false hypotheses are usually scattered across the reference map, they cannot form a stable local geographic consensus with other candidates. As a result, they receive little or no consensus reward, and their final scores remain close to their base scores.

In contrast, true geographic hypotheses tend to be spatially clustered. For example, in the rural scene shown in Fig.~\ref{Quantitative_vis}(b), the correct candidates in Row 2 and Row 3 are located in adjacent geographic regions. Through CD-RAPS, these candidates mutually reinforce each other by accumulating geographic consensus votes, which increases their final scores beyond the original base scores, e.g., from 0.898 to 1.057. This spatial voting mechanism helps distinguish geographically consistent hypotheses from isolated visual decoys, enabling SCC-Loc to select the correct location and achieve low prediction errors of 1.910~m and 4.306~m in the illustrated cases.
}

\subsection{Comparison with Two-Stage Baselines}

\textcolor{black}{To comprehensively evaluate the effectiveness of SCC-Loc, we compare it with systematically constructed two-stage baselines under the Top-3, Top-5, and Top-10 candidate settings, as summarized in Table~\ref{tab:localization_results}. In this comparison, Recall@$N$ is used to evaluate the coverage of the coarse retrieval stage, denoting the proportion of queries whose ground-truth location falls within the Top-$N$ retrieved satellite candidates. In contrast, Acc@5/10/20 measures the final localization accuracy under metric error thresholds.} Based on these metrics, the quantitative results reveal several critical insights regarding the cross-modal geo-localization bottleneck.

{\color{black}
\textbf{1) Modality Gap in Coarse Retrieval:}
Directly applying visible-light retrieval models to thermal imagery leads to severe performance degradation. For example, CAMP-based baselines achieve only 75.15\% Recall@3, reflecting the substantial domain shift between thermal UAV queries and visible satellite references. In contrast, visual foundation models such as DINOv2 and DINOv3 exhibit stronger zero-shot cross-modal generalization, increasing Recall@10 to over 99.45\%. This indicates that high-level foundation features can provide more modality-robust semantic representations for coarse thermal-visible retrieval.}

% \textbf{2) The Fragility of Conventional Matchers:} 
% Precise geometric solving exposes the vulnerability of disjointed pipelines that rely solely on inlier counts. Confronted with repetitive topological visual decoys, incorrect candidates often yield deceptively high inliers, causing uniform matcher failure. Specifically, at Top-10, the intra-modal RoMa yields a mere 5.02\% Acc@5 (with DINOv3), the cross-modal XoFTR suffers extreme geometric degradation ($95.59 \pm 177.26$ m mean error), and even the domain-adapted MINIMA$_{\text{RoMa}}$ plateaus at 6.64\%. This proves that implicitly trusting raw dense correspondences traps systems in a \textcolor{black}{``low residual, high drift''} state.

{\color{black}
\textbf{2) Fragility of Conventional Matchers:}
Precise geometric solving reveals the inherent fragility of disjointed pipelines that select candidate poses mainly according to raw matching scores or inlier counts. Under repetitive topological patterns, incorrect satellite candidates can still generate geometrically plausible correspondences and deceptively high inlier counts, causing conventional matchers to fail consistently. For example, under the Top-10 setting, RoMa achieves only 5.02\% Acc@5 with DINOv3, XoFTR suffers from severe geometric degradation with a mean error of $95.59 \pm 177.26$~m, and the domain-adapted MINIMA$_{\text{RoMa}}$ still plateaus at 6.64\% Acc@5. This indicates that relying on raw dense correspondences alone is insufficient for thermal-visible geo-localization and may lead to the \textcolor{black}{``low residual, high drift''} failure mode.}

% \textbf{3) Effectiveness of the SCC-Loc Framework:}
% SCC-Loc \textcolor{black}{effectively addresses} this bottleneck. By integrating semantic viewport alignment, cascaded filtering, and multi-dimensional reliability evaluation, we suppress the mean error to single digits ($9.37 \pm 25.26$ m at Top-10), while achieving a 7.6$\times$ improvement in Acc@5 over the strongest Acc@5 baseline under the strict 5-meter threshold (DINOv2 + MINIMA$_{\text{RoMa}}$). \textcolor{black}{
% Furthermore, the Top-3 to Top-10 improvement is jointly driven by increased retrieval coverage, C-SATSF-based correspondence purification, and CD-RAPS-based pose selection. As Recall@$N$ rises with more candidates, the candidate pool contains more true-positive satellite regions but also more visual decoys. C-SATSF first improves the geometric quality of each candidate by suppressing low-texture ambiguities and structurally inconsistent outliers, producing more reliable and spatially balanced correspondences for pose optimization. Built upon these purified correspondences, CD-RAPS exploits the high-recall yet noisy pool through physically constrained non-linear pose optimization and geographic consensus, raising SCC-Loc's Acc@5 from 33.49\% to 52.09\%, whereas baselines without such purification-and-selection mechanisms stagnate or degrade.
% }
\textbf{3) Effectiveness of the SCC-Loc Framework:}
SCC-Loc \textcolor{black}{effectively addresses} this bottleneck. By integrating semantic viewport alignment, cascaded filtering, and multi-dimensional reliability evaluation, we suppress the mean error to single digits ($9.37 \pm 25.26$ m at Top-10), while achieving a 7.6$\times$ improvement in Acc@5 over the strongest Acc@5 baseline under the strict 5-meter threshold (DINOv2 + MINIMA$_{\text{RoMa}}$). \textcolor{black}{
Furthermore, the Top-3 to Top-10 improvement is jointly driven by increased retrieval coverage, C-SATSF-based correspondence purification, and CD-RAPS-based pose selection. As Recall@$N$ rises, more true-positive regions are retrieved together with more visual decoys. C-SATSF improves candidate-level correspondence quality by suppressing ambiguous and structurally inconsistent matches, while CD-RAPS further selects the optimal pose through physically constrained optimization and geographic consensus. This purification-and-selection synergy raises SCC-Loc's Acc@5 from 33.49\% to 52.09\%, whereas baselines without such mechanisms stagnate or degrade.
}

\textcolor{black}{
\textbf{4) Precision-Latency Trade-off:}
Regarding computational efficiency, lightweight disjointed combinations such as DINOv2 + XoFTR execute rapidly, requiring only 0.92~s/query under the Top-3 setting, but suffer from severe accuracy degradation (1.66\% Acc@5). Although SCC-Loc incurs slightly higher latency than the strongest disjointed baseline under the Top-10 setting (14.21~s vs. 12.28~s), the additional cost is mainly devoted to cascaded correspondence purification and physically constrained pose verification. This design improves geometric reliability by suppressing structural outliers and visual decoys, leading to a more effective precision-latency trade-off for robust thermal-visible UAV geo-localization.}

\subsection{Comparison against Domain-Specific Baselines}

To further evaluate the effectiveness of the proposed SCC-Loc, we compare it with domain-specific networks STHN and NIVnet by expanding the search area from $600\times600$ to $1200\times1200$ m. The Top-10 retrieval results are summarized in Table~\ref{tab:separate_comparison}.

\textbf{1) Brittleness of End-to-End Regression:} 
Despite supervised training on Thermal-UAV, both STHN and NIVnet exhibit performance collapse in large-scale search spaces, with Acc@5 consistently $\le 1.00\%$. This failure stems from their reliance on direct parameter regression without hierarchical search, making them highly susceptible to initial viewpoint discrepancies and visual decoys in unconstrained global localization.

\textbf{2) Robustness Against Scale Variations:} 
As the search area quadruples ($600\times600$ to $1200\times1200$ m), the strongest baseline (DINOv2 + MINIMA$_{\text{RoMa}}$) suffers severe degradation. In contrast, SCC-Loc demonstrates superior scalability. Under the $1200\times1200$ m setting, it maintains a robust 46.30\% Acc@5, alongside high Acc@10 (76.00\%) and Acc@20 (83.45\%). Although the mean error increases to 46.62 m due to positional drifts from a few unlocalized queries, the consistently high Acc@10 and Acc@20 values indicate that our proposed method remains effective in large-scale search spaces.

\textbf{3) Deployment Efficiency and Memory Advantage:} 
While end-to-end models offer rapid inference ($\le 0.04$ s), their accuracy is insufficient for reliable deployment. Among viable two-stage strategies, SCC-Loc presents a highly advantageous trade-off. Although our pipeline requires slightly higher latency (14.21 s compared to 12.28 s) for \textcolor{black}{cascaded} filtering and optimization, its unified architecture yields significant memory savings. By sharing a single-load DINOv2 backbone across both retrieval and matching stages, SCC-Loc restricts peak GPU memory to 1174.7 MB—effectively halving the 2335.5 MB overhead of disjointed baselines. This balance of geometric precision and resource efficiency is critical for deployment on constrained UAV platforms.

\begin{figure*}[!t]
    \centering
    \includegraphics[width=\linewidth]{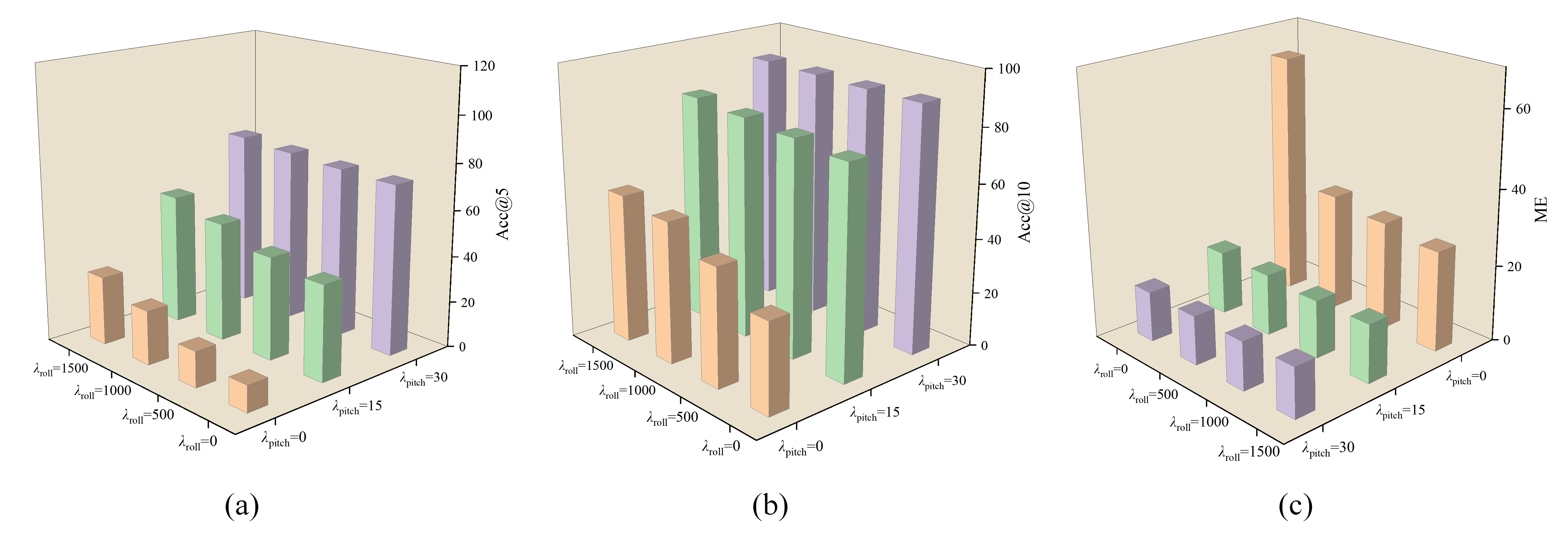}
    \caption{\textcolor{black}{Sensitivity analysis of the roll and pitch penalty weights in physically constrained non-linear pose optimization under the Top-5 retrieval setting. (a) Acc@5, (b) Acc@10, and (c) mean error (ME) under different combinations of $\lambda_{\mathrm{roll}}$ and $\lambda_{\mathrm{pitch}}$.}}
    \label{vis_pose_weight_sensitivity}
\end{figure*}

\begin{table}[!t]
\centering
\caption{Ablation study of the proposed framework under the Top-5 retrieval setting. The best results are highlighted in \textbf{bold}.}
\label{tab:ablation_study}
\renewcommand{\arraystretch}{1.4} 
% 完美控制列距
\setlength{\tabcolsep}{1.4pt}       
\resizebox{\linewidth}{!}{
    \begin{tabular}{ccc c ccc} % 增加了一列 c
    \toprule
    % --- 表头设计：左边是组件，右边是指标 ---
    \multicolumn{3}{c}{\textbf{Components}} & \multicolumn{4}{c}{\textbf{Performance Metrics}} \\ % 跨度从3改为4
    \cmidrule(r){1-3} \cmidrule(l){4-7} % 线段延伸到第7列
    SGVA & C-SATSF & CD-RAPS & \textbf{Recall@5} & \textbf{Acc@5/10/20} & \textbf{ME $\pm$ SD} & \textbf{Time} \\ 
    \midrule
    
    % --- 第一组：单个模块 (Base) ---
    \checkmark & & & 96.98 & 5.91 / 23.49 / 56.38 & 33.74 $\pm$ 61.53 & 6.81 \\
     & \checkmark & & 96.72 & 7.15 / 21.87 / 53.40 & 37.40 $\pm$ 68.48 & 6.82 \\
     & & \checkmark & 96.72 & 17.40 / 46.47 / 74.72 & 26.08 $\pm$ 55.59 & 8.60 \\ 
    \midrule
    
    % --- 第二组：两两组合 (Synergy) ---
    \checkmark & \checkmark & & 96.98 & 8.00 / 24.47 / 57.70 & 32.73 $\pm$ 60.38 & 7.00 \\
    \checkmark & & \checkmark & 96.98 & 20.89 / 51.91 / 78.00 & 21.71 $\pm$ 46.61 & 8.58 \\
     & \checkmark & \checkmark & 96.72 & 49.66 / 78.13 / 87.32 & 18.99 $\pm$ 54.25 & 7.36 \\ 
    \midrule
    
    % --- 第三组：完整模型 (Full) ---
    \checkmark & \checkmark & \checkmark & \textbf{96.98} & \textbf{51.66} / \textbf{81.32} / \textbf{89.87} & \textbf{15.30 $\pm$ 44.54} & \textbf{7.52} \\ 
    \bottomrule
    \end{tabular}
}
\end{table}

\begin{table}[!t]
\centering
\caption{\textcolor{black}{Runtime breakdown of SCC-Loc under the Top-10 retrieval setting.}}
\label{tab:runtime_breakdown}
\footnotesize
\setlength{\tabcolsep}{9.5pt}
\renewcommand{\arraystretch}{1.53}
\begin{tabular}{l l c}
\toprule
\textbf{Stage} & \textbf{Description} & \textbf{Time (s)} \\
\midrule
Retrieval & Top-10 candidate search & 0.769 \\
SGVA & Semantic viewport alignment & 0.001 \\
Dense Matching & MINIMA$_{\text{RoMa}}$ matching & 11.917 \\
C-SATSF & Cascaded correspondence filtering & 0.087 \\
CD-RAPS & Pose optimization and selection & 1.355 \\
Other & Data conversion and I/O overhead & 0.081 \\
\midrule
Total & Complete SCC-Loc pipeline & 14.210 \\
\bottomrule
\end{tabular}
\end{table}

\subsection{Ablation Study}
To \textcolor{black}{evaluate} the individual and synergistic contributions of the proposed modules (SGVA, C-SATSF, and CD-RAPS) within the SCC-Loc framework, we conduct an ablation study under the Top-5 retrieval setting, with results detailed in Table~\ref{tab:ablation_study}.

\textbf{1) Impact of Consensus-Driven Selection:} 
CD-RAPS provides the most significant standalone improvement, achieving 17.40\% Acc@5 and reducing ME to 26.08 m—outperforming standalone SGVA (5.91\%) and C-SATSF (7.15\%). This confirms that in selecting the optimal candidate location, replacing naive inlier counting with a geographic consensus voting mechanism—guided by multi-dimensional reliability evaluation and physical priors—is the most critical intervention for mitigating massive positional drift in cross-modal scenarios.

\textbf{2) Synergy of Purification and Consensus:}
The coupling of C-SATSF and CD-RAPS yields a \textcolor{black}{substantial performance improvement}, increasing Acc@5 from 17.40\% to 49.66\% and further compressing ME to 18.99 m.
\textcolor{black}{This improvement is mainly driven by the purified correspondences produced by C-SATSF. Although fewer matches are retained, they are more reliable and spatially balanced, providing a stronger geometric basis for the physically constrained pose optimization in CD-RAPS. Consequently, the optimized poses and reliability scores become more discriminative, while geographic consensus serves as a secondary refinement for final selection.}
By filtering low-texture ambiguities and topological distractors while ensuring a uniform distribution of the retained matches, C-SATSF provides a purified geometric foundation that prevents the CD-RAPS voting mechanism from being overwhelmed by structurally similar outliers or biased by locally clustered points.

\textbf{3) Efficacy of Semantic Viewport Alignment:}
Building upon the synergistic foundation of C-SATSF and CD-RAPS, integrating SGVA completes the cohesive SCC-Loc architecture, achieving optimal overall performance. It not only minimizes the mean error to 15.30 m and yields a peak Acc@5 of 51.66\%, but also maintains consistent improvements across all accuracy thresholds. Furthermore, it sustains the highest coarse retrieval recall (96.98\%). SGVA effectively resolves spatial quantization bias inherent in grid-based database construction. By adaptively aligning the reference viewport with the UAV's semantic field-of-view, it prevents target marginalization. This alignment allows the subsequent matching and optimization stages to operate on \textcolor{black}{better-centered satellite patches}, thereby \textcolor{black}{improving the robustness of the coarse-to-fine geo-localization pipeline}.

\textbf{4) Computational Efficiency:}
Standalone CD-RAPS incurs the highest latency (8.60 s) due to its iterative physics-constrained pose optimization. However, coupling it with C-SATSF reduces this time to 7.36 s. By filtering ambiguous outliers beforehand, C-SATSF provides a purified correspondence set that significantly accelerates the constrained solver's convergence. \textcolor{black}{Ultimately, under the Top-5 ablation setting, the full SCC-Loc pipeline requires 7.52 s per query,} achieving a favorable trade-off between localization accuracy and computational efficiency.

% \textcolor{black}{
% To further clarify the runtime of the complete Top-10 task, we provide a stage-wise breakdown in Table~\ref{tab:runtime_breakdown}. The results show that the dominant cost comes from repeated dense cross-modal matching, which accounts for 11.917 s of the 14.210 s total runtime. In contrast, SGVA introduces negligible overhead because it only performs semantic heatmap-based viewport adjustment. C-SATSF requires 0.087 s for cascaded correspondence purification, while CD-RAPS takes 1.355 s for multi-candidate pose optimization and reliability-aware selection. This breakdown indicates that SCC-Loc's additional latency is mainly spent on robust dense matching and physically constrained pose verification, which are necessary for suppressing structural outliers and visual decoys under severe thermal-visible modality gaps.
% }

\textcolor{black}{
To further clarify the computational cost of each component, Table~\ref{tab:runtime_breakdown} provides a stage-wise runtime breakdown of SCC-Loc under the Top-10 retrieval setting. The results show that the dominant cost comes from repeated dense cross-modal matching, which accounts for 11.917 s of the 14.210 s total runtime. In contrast, SGVA introduces negligible overhead because it only performs semantic heatmap-based viewport adjustment. C-SATSF requires 0.087 s for cascaded correspondence purification, while CD-RAPS takes 1.355 s for multi-candidate pose optimization and reliability-aware selection. This breakdown indicates that SCC-Loc's additional latency is mainly spent on robust dense matching and physically constrained pose verification, which are necessary for suppressing structural outliers and visual decoys under severe thermal-visible modality gaps.
}
% \begin{table*}[!t]
% \centering
% \caption{\textcolor{black}{Sensitivity analysis of roll and pitch penalty weights in non-linear pose optimization.}}
% \label{tab:pose_weight_sensitivity}
% \small
% \setlength{\tabcolsep}{4pt}
% \renewcommand{\arraystretch}{1.15}
% \resizebox{\textwidth}{!}{
% \begin{tabular}{c cccc cccc cccc}
% \toprule
% \multirow{2}{*}{$\lambda_{\mathrm{roll}}$}
% & \multicolumn{4}{c}{$\lambda_{\mathrm{pitch}}=0$}
% & \multicolumn{4}{c}{$\lambda_{\mathrm{pitch}}=15$}
% & \multicolumn{4}{c}{$\lambda_{\mathrm{pitch}}=30$} \\
% \cmidrule(lr){2-5} \cmidrule(lr){6-9} \cmidrule(lr){10-13}
% & \textbf{Acc@5} & \textbf{Acc@10} & \textbf{Acc@20} & \textbf{ME$\pm$SD}
% & \textbf{Acc@5} & \textbf{Acc@10} & \textbf{Acc@20} & \textbf{ME$\pm$SD}
% & \textbf{Acc@5} & \textbf{Acc@10} & \textbf{Acc@20} & \textbf{ME$\pm$SD} \\
% \midrule
% 0
% & 11.45 & 33.28 & 60.60 & 63.58$\pm$123.77
% & 41.32 & 77.49 & 89.06 & 16.77$\pm$46.54
% & 73.70 & 90.38 & 93.32 & 13.29$\pm$46.02 \\

% 500
% & 15.53 & 43.57 & 71.87 & 30.75$\pm$68.88
% & 44.77 & 79.57 & 89.53 & 16.25$\pm$46.61
% & 74.09 & 90.30 & 93.66 & 13.03$\pm$46.14 \\

% 1000
% & 23.49 & 51.87 & 75.53 & 28.59$\pm$69.38
% & 51.66 & 81.32 & 89.87 & 15.30$\pm$44.54
% & 75.36 & 90.72 & 93.49 & \textbf{12.92$\pm$45.70} \\

% 1500
% & 29.74 & 54.38 & 75.87 & 26.38$\pm$64.17
% & 56.64 & 83.15 & 90.51 & 15.44$\pm$47.58
% & \textbf{76.81} & \textbf{91.11} & \textbf{93.49} & 13.27$\pm$47.47 \\
% \bottomrule
% \end{tabular}
% }
% \end{table*}

\subsection{\textcolor{black}{Parameter Sensitivity Analysis}}

\textcolor{black}{
To further justify the empirical hyperparameter settings of SCC-Loc, we conduct sensitivity analyses for the key parameters in SGVA and CD-RAPS. Specifically, we evaluate the influence of the SGVA parameters $(\alpha,\beta)$, the roll/pitch penalty weights in the non-linear pose optimization, and the reliability weights in CD-RAPS. All sensitivity experiments are conducted under the Top-5 retrieval setting unless otherwise specified. These analyses aim to verify whether the adopted parameters provide stable performance under moderate perturbations, rather than being highly dependent on a single manually selected configuration.
}

\textcolor{black}{
\textbf{1) Sensitivity of SGVA Parameters:}
Table~\ref{tab:sgva_parameter_sensitivity} reports the sensitivity of the SGVA parameters $\alpha$ and $\beta$, where $\alpha$ controls viewport expansion and $\beta$ controls semantic center-shift strength. The tested ranges are determined according to their geometric meanings and a coarse grid search around the default setting. The results show that introducing viewport expansion compared with $\alpha=0$ consistently improves Acc@5/10/20 and reduces the mean error, indicating that additional peripheral context benefits dense matching. Increasing $\beta$ generally further improves localization accuracy by better compensating for spatial quantization bias. The best result is obtained at $(\alpha,\beta)=(0.4,1.5)$, achieving 54.26\% Acc@5 and 13.45~m mean error.
}

\begin{table}[!t]
\centering
\caption{\textcolor{black}{Sensitivity analysis of SGVA parameters under the Top-5 retrieval setting. The best results are highlighted in \textbf{bold}.}}
\label{tab:sgva_parameter_sensitivity}
\renewcommand{\arraystretch}{1.1}
\setlength{\tabcolsep}{7pt}
\resizebox{\linewidth}{!}{%
\begin{tabular}{ccccc}
\toprule
\multicolumn{2}{c}{\textbf{SGVA Parameters}} & \multicolumn{3}{c}{\textbf{Performance Metrics}} \\
\cmidrule(r){1-2} \cmidrule(l){3-5}
$\alpha$ & $\beta$ & \textbf{Recall@5} & \textbf{Acc@5/10/20} & \textbf{ME $\pm$ SD} \\
\midrule

\multirow{3}{*}{0}
  & 1.0 & 96.89 & 50.47 / 79.32 / 87.62 & 16.99 $\pm$ 48.01 \\
  & 1.2 & 96.94 & 49.66 / 77.79 / 87.83 & 17.80 $\pm$ 50.78 \\
  & 1.5 & 96.98 & 51.49 / 79.23 / 87.36 & 16.54 $\pm$ 46.29 \\
\cmidrule{2-5}

\multirow{3}{*}{0.2}
  & 1.0 & 96.98 & 51.66 / 81.32 / 89.87 & 15.30 $\pm$ 44.54 \\
  & 1.2 & 97.11 & 52.00 / 82.60 / 89.91 & 14.75 $\pm$ 43.66 \\
  & 1.5 & 96.59 & 52.15 / 81.44 / 89.53 & 15.55 $\pm$ 45.50 \\
\cmidrule{2-5}

\multirow{3}{*}{0.4}
  & 1.0 & 97.57 & 53.66 / 83.19 / 90.64 & 14.55 $\pm$ 43.83 \\
  & 1.2 & 97.62 & 53.11 / 83.79 / \textbf{90.77} & 13.48 $\pm$ 39.48 \\
  & 1.5 & \textbf{97.62} & \textbf{54.26} / \textbf{84.09} / 90.72 & \textbf{13.45 $\pm$ 39.27} \\
\bottomrule
\end{tabular}%
}
\end{table}

\begin{table}[b]
\centering
\caption{\textcolor{black}{Sensitivity analysis of CD-RAPS reliability weights under the Top-5 setting. The best results are highlighted in \textbf{bold}.}}
\label{tab:cdraps_weight_sensitivity}
% \scriptsize
\setlength{\tabcolsep}{5pt}
\renewcommand{\arraystretch}{1.3}
\begin{tabular}{c c c c c}
\toprule
$(w_1,w_2,w_3,w_4)$ & \textbf{Acc@5} & \textbf{Acc@10} & \textbf{Acc@20} & \textbf{ME$\pm$SD} \\
\midrule
(0.25, 0.25, 0.25, 0.25) & 50.17 & 80.60 & 89.23 & 15.27$\pm$42.93 \\
(0.1, 0.3, 0.3, 0.3)    & 50.68 & 81.40 & 89.74 & \textbf{14.90$\pm$43.07} \\
(0.1, 0.2, 0.35, 0.35)  & 51.66 & 81.32 & 89.87 & 15.30$\pm$44.54 \\
(0.2, 0.2, 0.3, 0.3)    & 51.15 & 81.62 & 89.45 & 14.97$\pm$42.52 \\
(0.1, 0.1, 0.4, 0.4)    & \textbf{53.40} & \textbf{82.81} & \textbf{90.00} & 15.12$\pm$44.55 \\
\bottomrule
\end{tabular}
\end{table}

% \textcolor{black}{
% \textbf{2) Sensitivity of Pose Penalty Weights:}
% Fig.~\ref{vis_pose_weight_sensitivity} illustrates the sensitivity of the roll and pitch penalty weights in the physically constrained non-linear pose optimization. As shown in Fig.~\ref{vis_pose_weight_sensitivity}(a) and (b), introducing pitch regularization significantly improves Acc@5 and Acc@10 compared with $\lambda_{\mathrm{pitch}}=0$, while Fig.~\ref{vis_pose_weight_sensitivity}(c) shows a clear reduction in mean error. For instance, when $\lambda_{\mathrm{roll}}=1000$, increasing $\lambda_{\mathrm{pitch}}$ from 0 to 15 improves Acc@5 from 23.49\% to 51.66\% and reduces ME from 28.59~m to 15.30~m, confirming that the onboard pitch prior provides an effective physical guide for pose convergence. Increasing either $\lambda_{\mathrm{pitch}}$ or $\lambda_{\mathrm{roll}}$ generally leads to better localization performance, indicating that stronger attitude constraints help suppress physically implausible pose solutions. Therefore, we adopt $(\lambda_{\mathrm{roll}},\lambda_{\mathrm{pitch}})=(1000,15)$ as a conservative setting that balances localization robustness and dependence on attitude priors. Although the exact values may be adjusted for UAV platforms with different IMU or gimbal stabilization accuracy, the consistent trends in Fig.~\ref{vis_pose_weight_sensitivity} suggest that the proposed physical regularization is not highly dependent on a single platform-specific configuration.}
\textcolor{black}{
\textbf{2) Sensitivity of Pose Penalty Weights:}
Fig.~\ref{vis_pose_weight_sensitivity} illustrates the sensitivity of the roll and pitch penalty weights in the physically constrained non-linear pose optimization. As shown in Fig.~\ref{vis_pose_weight_sensitivity}(a)--(c), introducing pitch regularization significantly improves Acc@5/10 and reduces ME. For instance, when $\lambda_{\mathrm{roll}}=1000$, increasing $\lambda_{\mathrm{pitch}}$ from 0 to 15 improves Acc@5 from 23.49\% to 51.66\% and reduces ME from 28.59~m to 15.30~m. This indicates that the pitch prior helps stabilize PnP optimization when DSM-derived elevation priors and cross-modal correspondences are noisy. Under near-nadir UAV observation, pitch is coupled with scale and horizontal translation; therefore, a stronger pitch penalty can suppress implausible attitude compensation and improve pose plausibility. We thus adopt $(\lambda_{\mathrm{roll}},\lambda_{\mathrm{pitch}})=(1000,15)$ as a conservative setting that balances localization robustness and dependence on attitude priors. The consistent trends suggest that this physical regularization is not overly sensitive to the exact weight selection and maintains stable performance within a reasonable parameter range.}

\textcolor{black}{
\textbf{3) Sensitivity of Reliability Weight Allocation:}
Table~\ref{tab:cdraps_weight_sensitivity} analyzes the reliability weight allocation in CD-RAPS. Compared with uniform weighting $(0.25,0.25,0.25,0.25)$, assigning larger weights to geometry-related indicators generally improves the final localization accuracy. In particular, increasing the weights of reprojection consistency and geometric uncertainty yields higher Acc@5/10/20, confirming that pose-level geometric reliability is more discriminative than semantic similarity or inlier count alone when visual decoys are present. The setting $(0.1,0.1,0.4,0.4)$ achieves the highest Acc@5, Acc@10, and Acc@20 in this sensitivity test, while the adopted default setting $(0.1,0.2,0.35,0.35)$ remains close to the best-performing configuration and retains a slightly larger contribution from the post-optimization inlier count for a more balanced reliability assessment. These results indicate that CD-RAPS is primarily driven by geometric reliability cues, while moderate variations in the reliability weights do not change the overall performance trend.
}

\begin{table}[!t]
\centering
\caption{\textcolor{black}{Robustness evaluation of the proposed framework under varying pitch and yaw angles in the Top-5 retrieval setting.}}
\label{tab:viewpoint_robustness}
\renewcommand{\arraystretch}{1.0553333} 
\setlength{\tabcolsep}{8pt} % 列数变少，稍微放宽一点列距，让排版更舒展
\resizebox{\linewidth}{!}{%
\begin{tabular}{ccccc} 
\toprule
% 性能指标恢复为 3 列，保持极简感
\multicolumn{2}{c}{\textbf{Noise Config}} & \multicolumn{3}{c}{\textbf{Performance Metrics}} \\
\cmidrule(r){1-2} \cmidrule(l){3-5} 
$\delta_{\text{pitch}}$ & $\delta_{\text{yaw}}$ & \textbf{Recall@5} & \textbf{Acc@5/10/20} & \textbf{ME $\pm$ SD} \\
\midrule

\multirow{3}{*}{0} 
  & 0  & 96.98 & 51.66 / 81.32 / 89.87 & 15.30 $\pm$ 44.54 \\
  & 10 & 97.36 & 51.02 / 82.00 / 89.70 & 15.42 $\pm$ 44.24 \\
  & 20 & 97.36 & 52.34 / 81.91 / 89.06 & 15.48 $\pm$ 43.87 \\
\cmidrule{2-5} % 局部横线

\multirow{3}{*}{10} 
  & 0  & 96.72 & 28.98 / 51.74 / 71.45 & 21.83 $\pm$ 44.71 \\
  & 10 & 97.19 & 28.17 / 51.66 / 70.51 & 22.31 $\pm$ 44.87 \\
  & 20 & 97.45 & 29.74 / 52.34 / 70.47 & 22.16 $\pm$ 44.70 \\
\cmidrule{2-5} % 局部横线

\multirow{3}{*}{20} 
  & 0  & 95.79 & 27.49 / 46.68 / 60.09 & 26.99 $\pm$ 45.67 \\
  & 10 & 96.55 & 26.55 / 46.26 / 58.81 & 28.61 $\pm$ 48.92 \\
  & 20 & 96.94 & 28.13 / 46.00 / 58.94 & 28.86 $\pm$ 51.79 \\
\bottomrule
\end{tabular}%
}
\end{table}

\subsection{Robustness Analysis}
To evaluate SCC-Loc against telemetry deviations induced by sensor inaccuracies or wind disturbances, we inject uniformly distributed noise into the pitch ($\delta_{\text{pitch}}$) and yaw ($\delta_{\text{yaw}}$) priors. Table~\ref{tab:viewpoint_robustness} details the Top-5 retrieval performance under varying noise configurations.

\textbf{1) Resilience to Yaw Perturbations:}
Although SCC-Loc utilizes a yaw prior for initial North-alignment, it demonstrates \textcolor{black}{strong robustness against yaw-prior deviations}. Evaluated with injected yaw noise ($\delta_{\text{yaw}}$) up to $20^{\circ}$, performance variations remain marginal across all pitch configurations, with Acc@R and mean errors maintaining \textcolor{black}{high stability with respect to yaw perturbations}. This confirms that our DINOv2-based cascaded pipeline possesses sufficient rotational invariance to tolerate biased yaw priors, supporting stable feature association and geometric solving under residual in-plane rotations.

\textbf{2) Sensitivity to Pitch Distortions:} 
Unlike yaw, pitch perturbations lead to a predictable performance decrease. As $\delta_{\text{pitch}}$ reaches $20^{\circ}$, Acc@5 drops to 28.13\% and mean error rises to 28.86 m. This sensitivity stems from the dual role of the pitch prior in our framework. First, pitch is used to estimate the effective GSD for satellite patch cropping; biased priors cause cross-modal scale inconsistencies and viewport misalignments. Second, the CD-RAPS strategy employs the pitch prior as a physical penalty term in non-linear pose optimization. \textcolor{black}{Large pitch noise biases the objective function, leading the optimization toward geometrically biased configurations and amplifying position error.}

Nevertheless, this represents a bounded degradation rather than systemic failure. Even under extreme dual-noise ($20^{\circ}$ pitch and $20^{\circ}$ yaw), SCC-Loc bounds the mean error to under 30 m—\textcolor{black}{still remaining more accurate than conventional baselines under the corresponding Top-5 setting.}

\section{Discussion} \label{Discussion}

Our analysis indicates that the thermal-visible modality gap manifests as a \textbf{structural illusion}, where geographically distinct entities—such as repetitive rooftops or identical agricultural plots—exhibit near-identical thermal signatures. This inherent ambiguity frequently deceives local matchers into generating dense but spatially erroneous correspondences. Consequently, relying solely on correspondence cardinality for pose identification is inherently unreliable, as false candidates often accumulate deceptively high inlier counts. Our findings indicate that integrating explicit physical constraints, multi-dimensional reliability evaluation, and geographic consensus is \textcolor{black}{important for distinguishing reliable geographic hypotheses from visual decoys}.

Despite these gains, SCC-Loc presents three primary limitations:
\begin{itemize}
    \item \textbf{Hyperparameter Dependency:} The framework relies on a relatively large number of manually defined hyperparameters, making the initialization process complex.
    \item \textbf{Telemetry Sensitivity:} Satellite image partitioning during retrieval and pose optimization are coupled with initial pitch and yaw priors; telemetry inaccuracies (e.g., gimbal jitter) can distort alignment and degrade precision.
    
    \item \textcolor{black}{\textbf{Inference Latency:} The complete pipeline currently requires approximately 14 s for single-pass inference. This latency mainly comes from the sequential execution of dense matching, cascaded match filtering, and non-linear pose optimization, which limits its direct deployment on resource-constrained onboard UAV platforms.}
\end{itemize}

\textcolor{black}{
These bottlenecks suggest a future transition from the current dense matching pipeline toward a lightweight learnable localization framework based on rotation-aware feature template matching. Specifically, DINOv2 can serve as a teacher model to distill a compact cross-modal feature extractor for UAV thermal imagery and satellite maps, enabling efficient generation of modality-robust feature maps. Instead of establishing dense point-wise correspondences, localization can then be reformulated as a feature-level template matching problem: the thermal UAV feature map is rotated over candidate yaw angles and matched against the satellite feature map to jointly estimate the horizontal position and heading. A coarse-to-fine search strategy can first locate high-probability regions globally and then refine the position and yaw angle within local satellite crops. Since correlation-based template matching can be efficiently accelerated via Fourier-domain convolution~\cite{yoo2009fast}, the future framework is expected to reduce the runtime of dense matching, cascaded filtering, and pose optimization, thereby improving onboard deployment efficiency.
}

\section{Conclusion} \label{Conclusion}
\textcolor{black}{In this paper, we propose SCC-Loc, a unified Semantic-Cascade-Consensus framework for UAV cross-modal Thermal Geo-localization in GNSS-denied environments. By sharing a \textcolor{black}{single foundation-model backbone} across retrieval and matching, SCC-Loc \textcolor{black}{provides a memory-efficient solution for handling} the thermal-visible modality gap. To overcome spatial misalignments, structural outliers, and decoy-induced positional drift, we integrate the SGVA module, C-SATSF mechanism, and CD-RAPS strategy. We also construct the Thermal-UAV dataset, which provides multi-temporal thermal queries, complex urban-rural scenes, and aligned 3D priors for cross-modal benchmarking. Nevertheless, SCC-Loc still relies on manually configured hyperparameters and onboard attitude priors, while dense matching and multi-candidate optimization introduce additional latency for onboard deployment. Future work will investigate adaptive parameter optimization, stronger tolerance to telemetry disturbances, and lightweight acceleration strategies toward real-time UAV localization.}

\bibliographystyle{IEEEtran}
\bibliography{sample-base}

@inproceedings{zheng2020university,
  title={University-1652: A multi-view multi-source benchmark for drone-based geo-localization},
  author={Zheng, Zhedong and Wei, Yunchao and Yang, Yi},
  booktitle={Proc. ACM Int. Conf. Multimedia},
  pages={1395--1403},
  year={2020}
}

@article{chen2024fpi,
  title={OS-FPI: A coarse-to-fine one-stream network for UAV geolocalization},
  author={Chen, Jiahao and Zheng, Enhui and Dai, Ming and Chen, Yifu and Lu, Yusheng},
  journal={IEEE J. Sel. Top. Appl. Earth Observ. Remote Sens.},
  volume={17},
  pages={7852--7866},
  year={2024},
  publisher={IEEE}
}

@article{couturier2024review,
  title={A review on deep learning for uav absolute visual localization},
  author={Couturier, Andy and Akhloufi, Moulay A},
  journal={Drones},
  volume={8},
  number={11},
  pages={622},
  year={2024},
  publisher={MDPI}
}

@article{avola2024uav,
  title={UAV Geo-Localization for Navigation: A Survey},
  author={Avola, Danilo and Cinque, Luigi and Emam, Emad and Fontana, Federico and Foresti, Gian Luca and Marini, Marco Raoul and Mecca, Alessio and Pannone, Daniele},
  journal={IEEE Access},
  year={2024},
  publisher={IEEE}
}

@article{dai2021transformer,
  title={A transformer-based feature segmentation and region alignment method for UAV-view geo-localization},
  author={Dai, Ming and Hu, Jianhong and Zhuang, Jiedong and Zheng, Enhui},
  journal={IEEE Trans. Circuits Syst. Video Technol.},
  volume={32},
  number={7},
  pages={4376--4389},
  year={2021},
  publisher={IEEE}
}

@article{keetha2023anyloc,
  title={Anyloc: Towards universal visual place recognition},
  author={Keetha, Nikhil and Mishra, Avneesh and Karhade, Jay and Jatavallabhula, Krishna Murthy and Scherer, Sebastian and Krishna, Madhava and Garg, Sourav},
  journal={IEEE Robot. Autom. Lett.},
  volume={9},
  number={2},
  pages={1286--1293},
  year={2023},
  publisher={IEEE}
}

@article{wu2025uav,
  title={UAV-GeoLoc: A Large-Vocabulary Dataset and Geometry-Transformed Method for UAV Geo-Localization},
  author={Wu, Rouwan and Deng, Jiacheng and Mou, Mingyu and He, Xingyi and Zhang, Maojun and Liu, Yu and Yan, Shen},
  journal={IEEE Robot. Autom. Lett.},
  year={2025},
  publisher={IEEE}
}

@article{xiao2024sthn,
  title={Sthn: Deep homography estimation for uav thermal geo-localization with satellite imagery},
  author={Xiao, Jiuhong and Zhang, Ning and Tortei, Daniel and Loianno, Giuseppe},
  journal={IEEE Robot. Autom. Lett.},
  year={2024},
  publisher={IEEE}
}

@article{meng2024airgeonet,
  title={AirGeoNet: A Map-Guided Visual Geo-Localization Approach for Aerial Vehicles},
  author={Meng, Xiangze and Guo, Wulong and Zhou, Kai and Sun, Ting and Deng, Lei and Yu, Shijie and Feng, Yuhao},
  journal={IEEE Trans. Geosci. Remote Sens.},
  year={2024},
  publisher={IEEE}
}

@article{ye2025exploring,
  title={Exploring the best way for UAV visual localization under Low-altitude Multi-view Observation Condition: a Benchmark},
  author={Ye, Yibin and Teng, Xichao and Chen, Shuo and Li, Zhang and Liu, Leqi and Yu, Qifeng and Tan, Tao},
  journal={arXiv:2503.10692},
  year={2025}
}

@article{li2025geovins,
  title={GeoVINS: Geographic-Visual-Inertial Navigation System for Large-Scale Drift-Free Aerial State Estimation},
  author={Li, Chunyu and He, Mengfan and Chen, Chao and Liu, Jiacheng and Lyu, Xu and Huang, Guoquan and Meng, Ziyang},
  journal={IEEE Trans. Robot.},
  year={2025},
  publisher={IEEE}
}

@inproceedings{zhang2016degeneracy,
  title={On degeneracy of optimization-based state estimation problems},
  author={Zhang, Ji and Kaess, Michael and Singh, Sanjiv},
  booktitle={Proc. IEEE Int. Conf. Robot. Autom. (ICRA)},
  pages={809--816},
  year={2016},
  organization={IEEE}
}

@inproceedings{xiao2023long,
  title={Long-range uav thermal geo-localization with satellite imagery},
  author={Xiao, Jiuhong and Tortei, Daniel and Roura, Eloy and Loianno, Giuseppe},
  booktitle={Proc. IEEE/RSJ Int. Conf. Intell. Robots Syst. (IROS)},
  pages={5820--5827},
  year={2023},
  organization={IEEE}
}

@article{yoo2009fast,
  title={Fast normalized cross-correlation},
  author={Yoo, Jae-Chern and Han, Tae Hee},
  journal={Circuits, Syst. Signal Process.},
  volume={28},
  number={6},
  pages={819--843},
  year={2009},
  publisher={Springer}
}

@inproceedings{detone2018superpoint,
  title={Superpoint: Self-supervised interest point detection and description},
  author={DeTone, Daniel and Malisiewicz, Tomasz and Rabinovich, Andrew},
  booktitle={Proc. IEEE Conf. Comput. Vis. Pattern Recognit. Workshops},
  pages={224--236},
  year={2018}
}

@inproceedings{sun2021loftr,
  title={LoFTR: Detector-free local feature matching with transformers},
  author={Sun, Jiaming and Shen, Zehong and Wang, Yuang and Bao, Hujun and Zhou, Xiaowei},
  booktitle={Proc. IEEE/CVF Conf. Comput. Vis. Pattern Recognit. (CVPR)},
  pages={8922--8931},
  year={2021}
}

@article{he2024leveraging,
  title={Leveraging map retrieval and alignment for robust UAV visual geo-localization},
  author={He, Mengfan and Liu, Jiacheng and Gu, Pengfei and Meng, Ziyang},
  journal={IEEE Trans. Instrum. Meas.},
  volume={73},
  pages={1--13},
  year={2024},
  publisher={IEEE}
}

@article{oquab2023dinov2,
  title={Dinov2: Learning robust visual features without supervision},
  author={Oquab, Maxime and Darcet, Timoth{\'e}e and Moutakanni, Th{\'e}o and Vo, Huy and Szafraniec, Marc and Khalidov, Vasil and Fernandez, Pierre and Haziza, Daniel and Massa, Francisco and El-Nouby, Alaaeldin and others},
  journal={arXiv:2304.07193},
  year={2023}
}

@inproceedings{edstedt2024roma,
  title={Roma: Robust dense feature matching},
  author={Edstedt, Johan and Sun, Qiyu and B{\"o}kman, Georg and Wadenb{\"a}ck, M{\aa}rten and Felsberg, Michael},
  booktitle={Proc. IEEE/CVF Conf. Comput. Vis. Pattern Recognit. (CVPR)},
  pages={19790--19800},
  year={2024}
}

@article{radenovic2018fine,
  title={Fine-tuning CNN image retrieval with no human annotation},
  author={Radenovi{\'c}, Filip and Tolias, Giorgos and Chum, Ond{\v{r}}ej},
  journal={IEEE Trans. Pattern Anal. Mach. Intell.},
  volume={41},
  number={7},
  pages={1655--1668},
  year={2018},
  publisher={IEEE}
}

@article{sola2018micro,
  title={A micro lie theory for state estimation in robotics},
  author={Sola, Joan and Deray, Jeremie and Atchuthan, Dinesh},
  journal={arXiv:1812.01537},
  year={2018}
}

@article{wu2024camp,
  title={Camp: A cross-view geo-localization method using contrastive attributes mining and position-aware partitioning},
  author={Wu, Qiong and Wan, Yi and Zheng, Zhi and Zhang, Yongjun and Wang, Guangshuai and Zhao, Zhenyang},
  journal={IEEE Trans. Geosci. Remote Sens.},
  year={2024},
  publisher={IEEE}
}

@article{dhaouadi2025ortholoc,
  title={OrthoLoC: UAV 6-DoF Localization and Calibration Using Orthographic Geodata},
  author={Dhaouadi, Oussema and Marin, Riccardo and Meier, Johannes and Kaiser, Jacques and Cremers, Daniel},
  journal={arXiv:2509.18350},
  year={2025}
}

@article{dosovitskiy2020image,
  title={An image is worth 16x16 words: Transformers for image recognition at scale},
  author={Dosovitskiy, Alexey},
  journal={arXiv:2010.11929},
  year={2020}
}

@article{wu2006pnp,
  title={PnP problem revisited},
  author={Wu, Yihong and Hu, Zhanyi},
  journal={J. Math. Imag. Vis.},
  volume={24},
  number={1},
  pages={131--141},
  year={2006},
  publisher={Springer}
}

@article{martinez2021search,
  title={Search and rescue operation using UAVs: A case study},
  author={Martinez-Alpiste, Ignacio and Golcarenarenji, Gelayol and Wang, Qi and Alcaraz-Calero, Jose Maria},
  journal={Expert Syst. Appl.},
  volume={178},
  pages={114937},
  year={2021},
  publisher={Elsevier}
}

@incollection{koslowski2021drones,
  title={Drones and border control: An examination of state and non-state actor use of UAVs along borders},
  author={Koslowski, Rey},
  booktitle={Research Handbook on International Migration and Digital Technology},
  pages={152--165},
  year={2021},
  publisher={Edward Elgar Publishing}
}

@inproceedings{xiao2025uasthn,
  title={Uasthn: Uncertainty-aware deep homography estimation for uav satellite-thermal geo-localization},
  author={Xiao, Jiuhong and Loianno, Giuseppe},
  booktitle={Proc. IEEE Int. Conf. Robot. Autom. (ICRA)},
  pages={14066--14072},
  year={2025},
  organization={IEEE}
}

@inproceedings{tuzcuouglu2024xoftr,
  title={Xoftr: Cross-modal feature matching transformer},
  author={Tuzcuo{\u{g}}lu, {\"O}nder and K{\"o}ksal, Aybora and Sofu, Bu{\u{g}}ra and Kalkan, Sinan and Alatan, A Aydin},
  booktitle={Proc. IEEE/CVF Conf. Comput. Vis. Pattern Recognit. (CVPR)},
  pages={4275--4286},
  year={2024}
}

@inproceedings{ren2025minima,
  title={Minima: Modality invariant image matching},
  author={Ren, Jiangwei and Jiang, Xingyu and Li, Zizhuo and Liang, Dingkang and Zhou, Xin and Bai, Xiang},
  booktitle={Proc. Comput. Vis. Pattern Recognit. Conf. (CVPR)},
  pages={23059--23068},
  year={2025}
}

@article{liu2024multimodal,
  title={Multimodal Absolute Visual Localization for Unmanned Aerial Vehicles},
  author={Liu, Zhunga and Li, Huandong and Zhang, Zuowei and Lyu, Yanyi and Xiong, Jiexuan},
  journal={IEEE Trans. Veh. Technol.},
  volume={73},
  number={11},
  pages={16402--16415},
  year={2024},
  publisher={IEEE}
}

@article{simeoni2025dinov3,
  title={Dinov3},
  author={Sim{\'e}oni, Oriane and Vo, Huy V and Seitzer, Maximilian and Baldassarre, Federico and Oquab, Maxime and Jose, Cijo and Khalidov, Vasil and Szafraniec, Marc and Yi, Seungeun and Ramamonjisoa, Micha{\"e}l and others},
  journal={arXiv:2508.10104},
  year={2025}
}

@article{xu2024uav,
  title={Uav-visloc: A large-scale dataset for uav visual localization},
  author={Xu, Wenjia and Yao, Yaxuan and Cao, Jiaqi and Wei, Zhiwei and Liu, Chunbo and Wang, Jiuniu and Peng, Mugen},
  journal={arXiv:2405.11936},
  year={2024}
}

@article{zhu2023sues,
  title={SUES-200: A multi-height multi-scene cross-view image benchmark across drone and satellite},
  author={Zhu, Runzhe and Yin, Ling and Yang, Mingze and Wu, Fei and Yang, Yuncheng and Hu, Wenbo},
  journal={IEEE Trans. Circuits Syst. Video Technol.},
  volume={33},
  number={9},
  pages={4825--4839},
  year={2023},
  publisher={IEEE}
}

@article{dai2023vision,
  title={Vision-based UAV self-positioning in low-altitude urban environments},
  author={Dai, Ming and Zheng, Enhui and Feng, Zhenhua and Qi, Lei and Zhuang, Jiedong and Yang, Wankou},
  journal={IEEE Trans. Image Process.},
  volume={33},
  pages={493--508},
  year={2023},
  publisher={IEEE}
}

@inproceedings{ji2025mmgeo,
  title={MMGeo: Multimodal Compositional Geo-Localization for UAVs},
  author={Ji, Yuxiang and He, Boyong and Tan, Zhuoyue and Wu, Liaoni},
  booktitle={Proc. IEEE/CVF Int. Conf. Comput. Vis. (ICCV)},
  pages={25165--25175},
  year={2025}
}

@inproceedings{arandjelovic2016netvlad,
  title={NetVLAD: CNN architecture for weakly supervised place recognition},
  author={Arandjelovic, Relja and Gronat, Petr and Torii, Akihiko and Pajdla, Tomas and Sivic, Josef},
  booktitle={Proc. IEEE Conf. Comput. Vis. Pattern Recognit. (CVPR)},
  pages={5297--5307},
  year={2016}
}

@article{vasile2025uav,
  title={UAV-TIRVis: A Benchmark Dataset for Thermal--Visible Image Registration from Aerial Platforms},
  author={Vasile, Costin-Emanuel and B{\^\i}r{\u{a}}, C{\u{a}}lin and Hobincu, Radu},
  journal={J. Imag.},
  volume={11},
  number={12},
  pages={432},
  year={2025},
  publisher={MDPI}
}

@article{liu2024mcgs,
  title={MCGS-ReID: A Visible-Infrared Vehicle Reidentification Method Using Modal-Cross Graph Sampler},
  author={Liu, Jianfei and Zhao, Chunhui and Zhao, Chen and Su, Nan and Lu, Wanxuan and Yan, Yiming and Feng, Shou and Qu, Yunfei},
  journal={IEEE J. Sel. Top. Appl. Earth Observ. Remote Sens.},
  volume={18},
  pages={18806--18818},
  year={2024},
  publisher={IEEE}
}

@article{xu2026enhancing,
  title={Enhancing UAV Geo-Location with Multi-Modal Transformer Networks: The MMGLT Approach},
  author={Xu, Wangping and Chen, Nanxing and Yuan, Jiayu and Fan, Jiqi and Chen, Weimin and Zheng, Enhui},
  journal={IEEE J. Sel. Top. Appl. Earth Observ. Remote Sens.},
  year={2026},
  publisher={IEEE}
}

@article{liu2024segcn,
  title={SeGCN: A semantic-aware graph convolutional network for UAV geo-localization},
  author={Liu, Xiangzeng and Wang, Ziyao and Wu, Yue and Miao, Qiguang},
  journal={IEEE J. Sel. Top. Appl. Earth Observ. Remote Sens.},
  volume={17},
  pages={6055--6066},
  year={2024},
  publisher={IEEE}
}

@article{ma2019locality,
  title={Locality preserving matching},
  author={Ma, Jiayi and Zhao, Ji and Jiang, Junjun and Zhou, Huabing and Guo, Xiaojie},
  journal={Int. J. Comput. Vis.},
  volume={127},
  number={5},
  pages={512--531},
  year={2019},
  publisher={Springer}
}

@article{jiang2019reliable,
  title={Reliable image matching via photometric and geometric constraints structured by Delaunay triangulation},
  author={Jiang, San and Jiang, Wanshou},
  journal={ISPRS J. Photogrammetry Remote Sens.},
  volume={153},
  pages={1--20},
  year={2019},
  publisher={Elsevier}
}

@article{lowry2015visual,
  title={Visual place recognition: A survey},
  author={Lowry, Stephanie and S{\"u}nderhauf, Niko and Newman, Paul and Leonard, John J and Cox, David and Corke, Peter and Milford, Michael J},
  journal={IEEE Trans. Robot.},
  volume={32},
  number={1},
  pages={1--19},
  year={2015},
  publisher={IEEE}
}

@article{qin2018vins,
  title={Vins-mono: A robust and versatile monocular visual-inertial state estimator},
  author={Qin, Tong and Li, Peiliang and Shen, Shaojie},
  journal={IEEE Trans. Robot.},
  volume={34},
  number={4},
  pages={1004--1020},
  year={2018},
  publisher={IEEE}
}

@misc{sengijpta1995fundamentals,
  title={Fundamentals of statistical signal processing: Estimation theory},
  author={Sengijpta, Sailes K},
  year={1995},
  publisher={Taylor \& Francis}
}

\vspace{-39pt}
\begin{IEEEbiography}[{\includegraphics[width=1in,height=1.25in,clip,keepaspectratio]{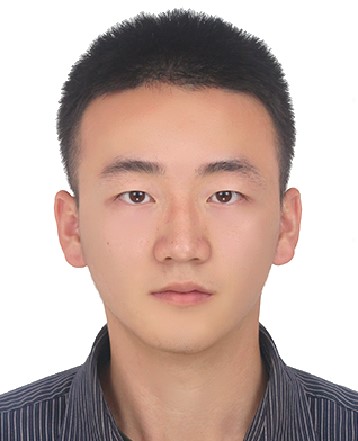}}]{Xiaoran Zhang}
received the B.E. degree in simulation engineering from the National University of Defense Technology, Changsha, China, in 2024, where he is currently pursuing the M.S. degree in control science and engineering. 

His research interests include unmanned systems, visual geo-localization and computer vision.
\end{IEEEbiography}

% ===================== 最后一页的 5 个作者 =====================

% 左栏第1个
\vspace{-42pt}
\begin{IEEEbiography}[{\includegraphics[width=1in,height=1.25in,clip,keepaspectratio]{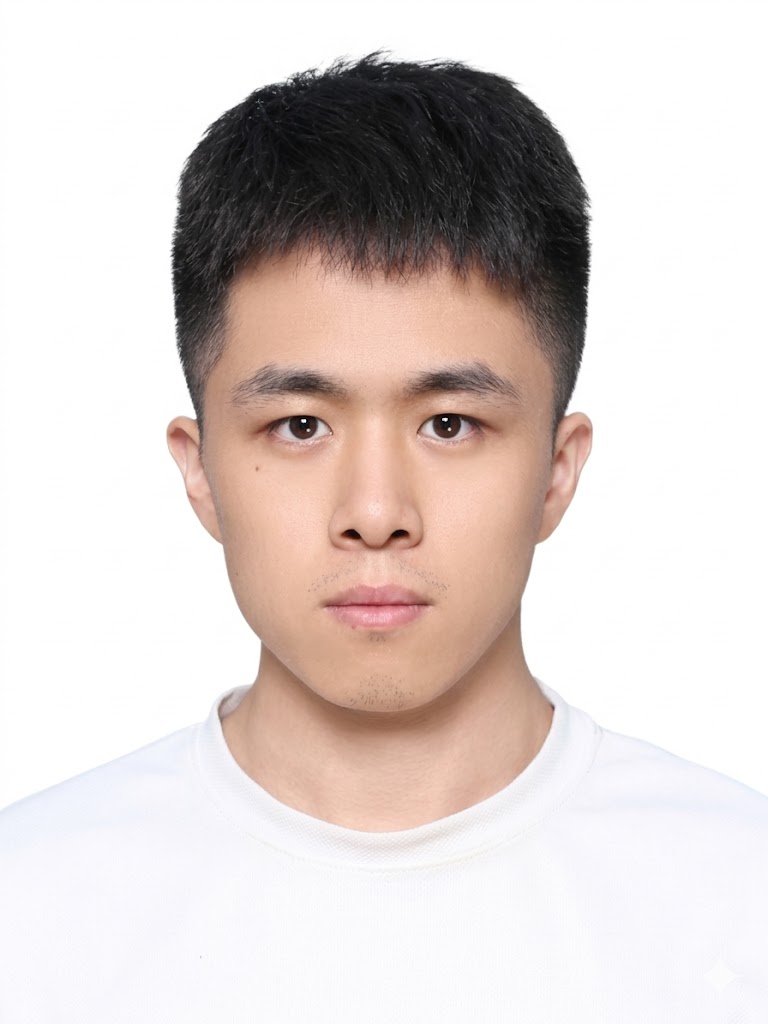}}]{Yu Liu}
received the B.E. degree in management science and engineering from the National University of Defense Technology, Changsha, China, in 2024, where he is currently pursuing the M.S. degree in control science and engineering. 

His research interests include visual geo-localization and computer vision.
\end{IEEEbiography}

% 左栏第2个
\vspace{-42pt}
\begin{IEEEbiography}[{\includegraphics[width=1in,height=1.25in,clip,keepaspectratio]{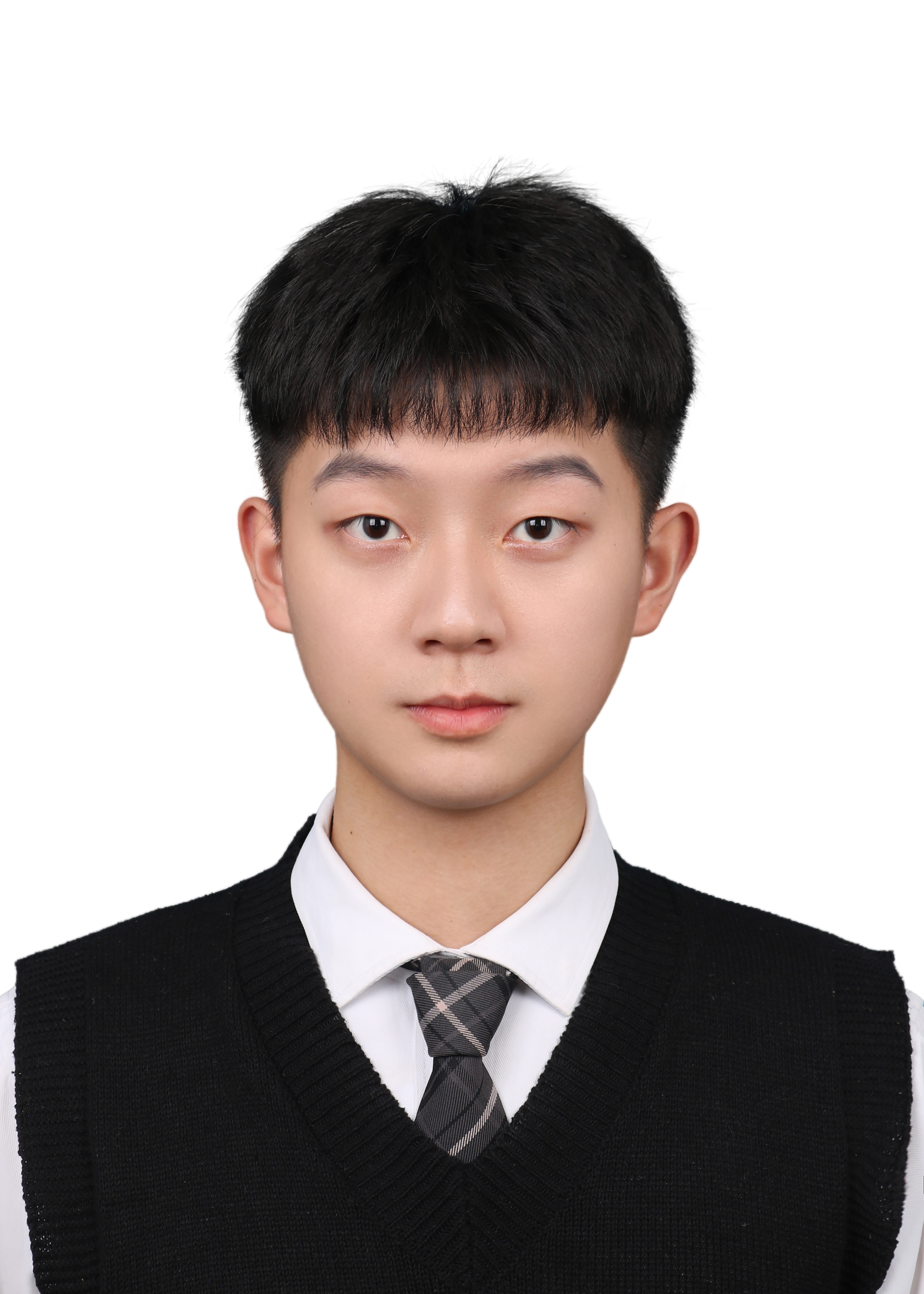}}]{Jinyu Liang}
received the B.E. degree in automation from the Central South University of Forestry and Technology, Changsha, China, in 2024, where he is currently pursuing the M.S. degree in control science and engineering. 

His research interests include image recognition and 3D reconstruction.
\end{IEEEbiography}

% 左栏第3个
\vspace{-42pt}
\begin{IEEEbiography}[{\includegraphics[width=1in,height=1.25in,clip,keepaspectratio]{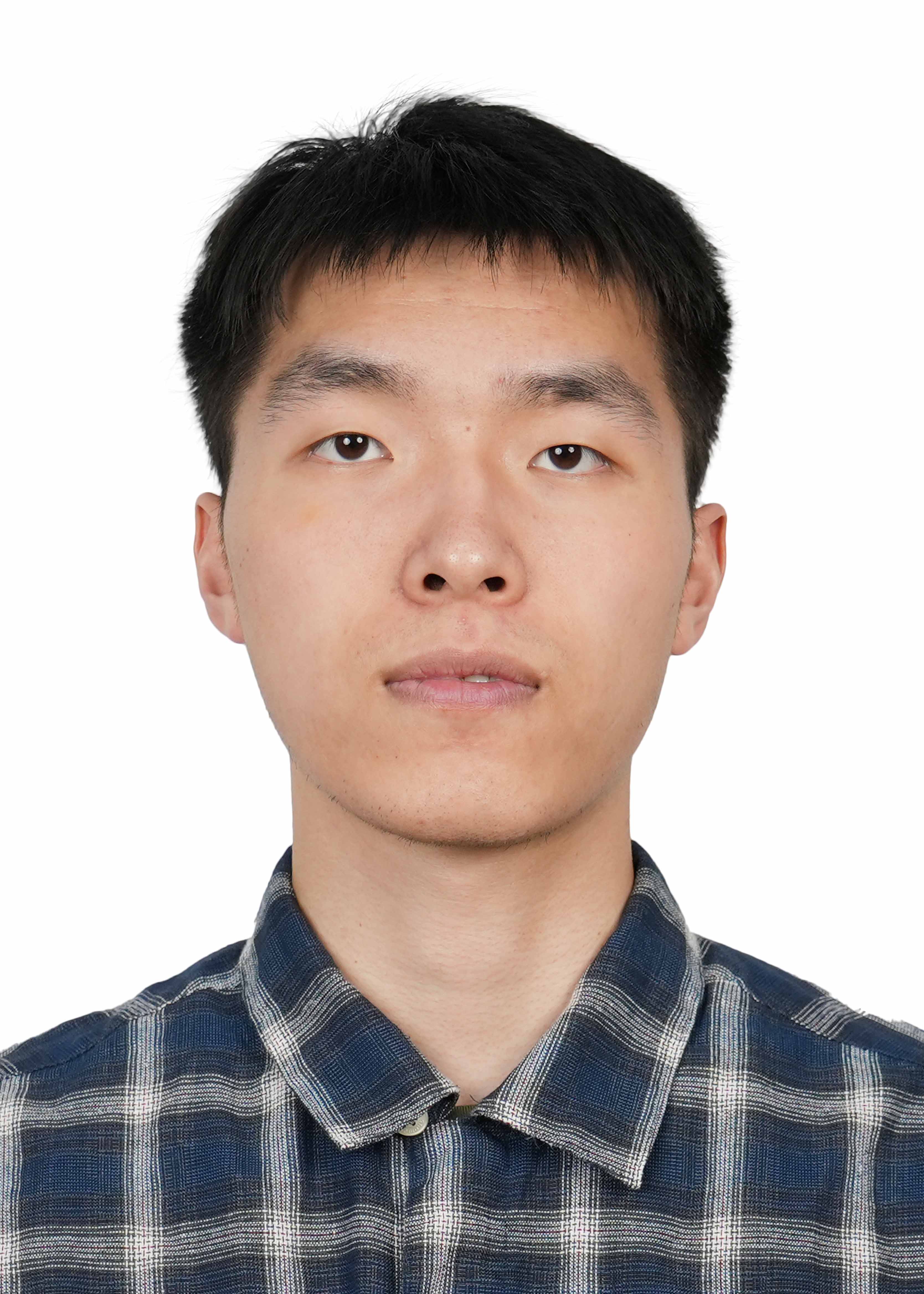}}]{Kangqiushi Li}
received the B.E. degree in simulation engineering from the National University of Defense Technology, Changsha, China, in 2024, where he is currently pursuing the M.S. degree in electronic information. 

His research interests include target detection and computer vision.
\end{IEEEbiography}

\vspace{-42pt}
\begin{IEEEbiography}[{\includegraphics[width=1in,height=1.25in,clip,keepaspectratio]{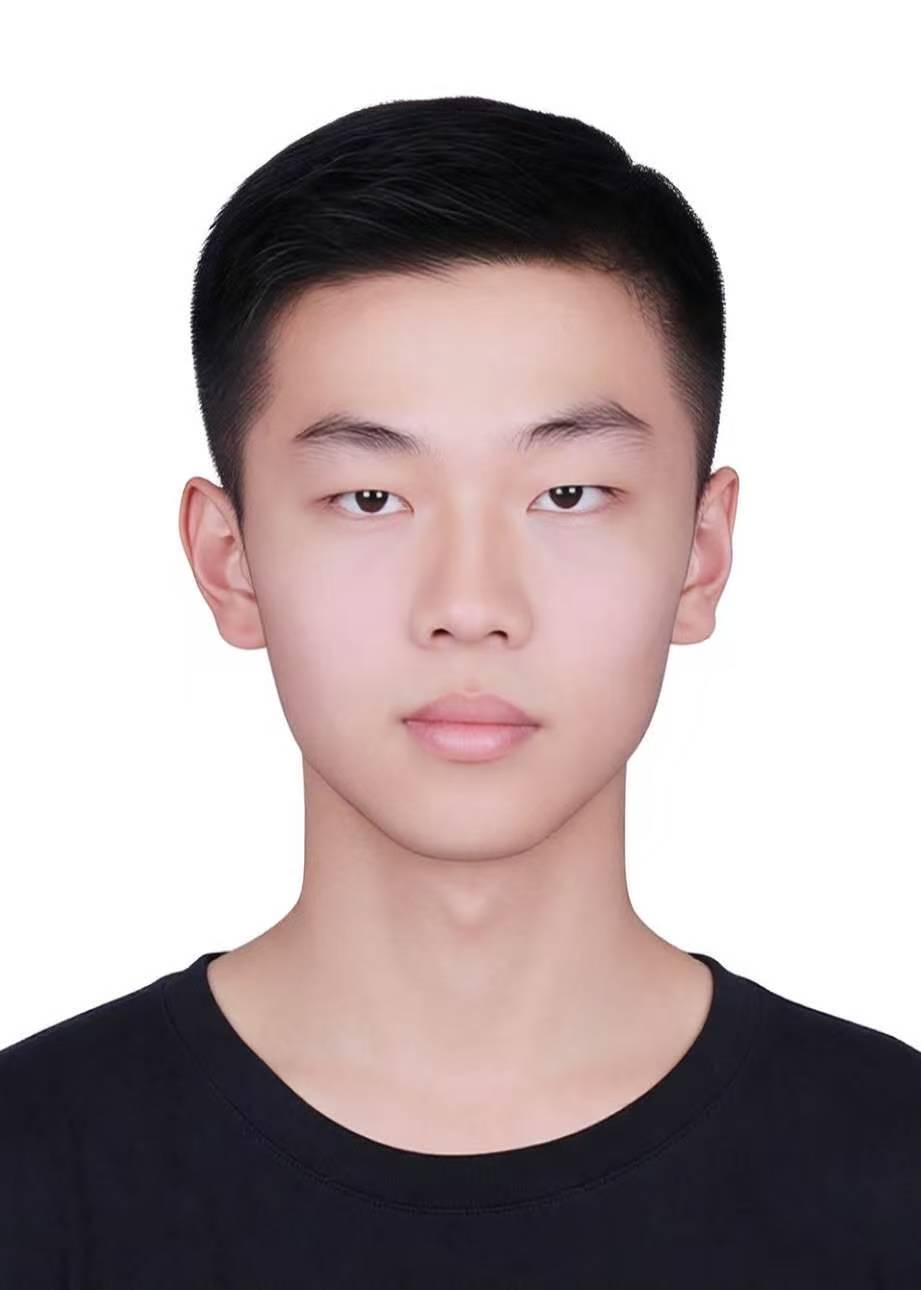}}]{Zhiwei Huang}
received the B.E. degree in management science and engineering from the National University of Defense Technology, Changsha, China, in 2025, where he is currently pursuing the M.S. degree in control science and engineering. 

His research interests include fast 3D reconstruction and computer vision.
\end{IEEEbiography}

\vspace{-42pt}
\begin{IEEEbiography}[{\includegraphics[width=1in,height=1.25in,clip,keepaspectratio]{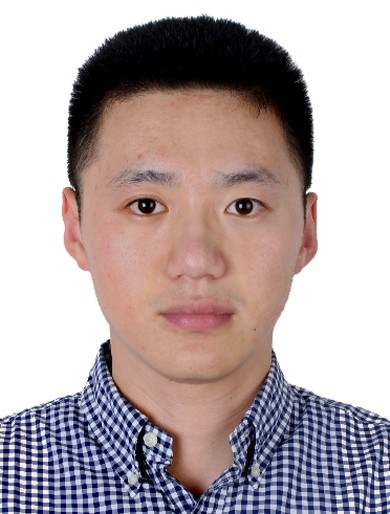}}]{Huaxin Xiao}
received the Ph.D. degree from the National University of Defense Technology, Changsha, China, in 2018. He is currently an Associate Professor with the College of System Engineering, National University of Defense Technology.

His current research interests include saliency detection and image/video object segmentation.
\end{IEEEbiography}

\end{document}